%% file: main.tex
\documentclass[10pt,twocolumn,letterpaper]{article}

\usepackage{iccv}
\usepackage{times}
\usepackage{epsfig}
\usepackage{graphicx}
\usepackage{amsmath}
\usepackage{amssymb}

\usepackage{pifont}
\newcommand{\cmark}{\ding{51}}%
\newcommand{\xmark}{\ding{55}}%

\usepackage{color}
\usepackage{algorithm}
\usepackage{algpseudocode}
\usepackage{dsfont}
\usepackage{subcaption}
\usepackage{cite}
\usepackage{booktabs}
\usepackage{tabularx}
\usepackage{xurl}
\usepackage{soul}
\usepackage{balance}
\usepackage[normalem]{ulem}

\usepackage{lscape} 
\usepackage{array,booktabs}

\usepackage[pagebackref=true,breaklinks=true,letterpaper=true,colorlinks,bookmarks=false]{hyperref}
\input{defs}

\iccvfinalcopy % *** Uncomment this line for the final submission

\ificcvfinal\fi

\begin{document}

%%%%%%%%% TITLE
\title{Spectral Graphormer: Spectral Graph-based Transformer for Egocentric Two-Hand Reconstruction using Multi-View Color Images}

\author{Tze Ho Elden Tse$^{1,2}$\thanks{This work was done during an internship at Google.} \: Franziska Mueller$^1$ \: Zhengyang Shen$^1$ \: Danhang Tang$^1$ \: Thabo Beeler$^1$ \\  Mingsong Dou$^1$ \: Yinda Zhang$^1$ \: Sasa Petrovic$^1$ \: Hyung Jin Chang$^2$ \: Jonathan Taylor$^1$ \: Bardia Doosti$^1$\\
$^1$Google \qquad $^2$University of Birmingham
}

\maketitle
\ificcvfinal\thispagestyle{empty}\fi

\input{tex/abstract}
\input{tex/intro}
\input{tex/related}

\input{tex/method}
\input{tex/result}
\input{tex/conclusion}

\vspace{-0.1cm}
% \section*{Acknowledgements}
\section*{\fontsize{10}{12}\selectfont Acknowledgements}
\vspace{-0.25cm}
\footnotesize \noindent This research was supported by the MSIT (Ministry of Science and ICT), Korea, under the ITRC (Information Technology Research Center) support program (IITP-2023-2020-0-01789), supervised by the IITP (Institute for Information \& Communications Technology Planning \& Evaluation).

\input{fig/qualitative/collision_full/item} 
\newpage
% \clearpage
{\small
\bibliographystyle{ieee_fullname}
\bibliography{ref}
}

{\normalsize
\appendix
\twocolumn[
\centering
\Large
\textbf{Spectral Graphormer: Spectral Graph-based Transformer for Egocentric Two-Hand Reconstruction using Multi-View Color Images} \\
\vspace{0.5em}Supplementary Material \\
\vspace{3.0em}
]
\input{tex/supplementary}
}
\end{document}

%% file: defs.tex
\newcommand{\comment}[1]{}

\newcommand{\bx}[0]{\mathbf{x}}

\newcommand{\bm}{\mathbf{m}}

\newcommand{\fig}[1]{Fig.~\ref{fig:#1}}

\newcommand{\loss}{\mathcal{L}}

\definecolor{orange}{rgb}{1,0.5,0}
\definecolor{blue}{rgb}{0,0,0.6}
\definecolor{green}{rgb}{0,0.6,0}
\definecolor{purple}{rgb}{0.5,0,0.5}

\definecolor{color1}{RGB}{0,199,1}
\definecolor{color2}{RGB}{224,43,28}

\newcolumntype{P}[1]{>{\centering\arraybackslash}p{#1}}

\renewcommand{\paragraph}[1]{\vspace{0.5em}\noindent\textbf{#1}}

%% file: tex/abstract.tex
\begin{abstract}
We propose a novel transformer-based framework that reconstructs two high fidelity hands from multi-view RGB images.
Unlike existing hand pose estimation methods, where one typically trains a deep network to regress hand model parameters from single RGB image, we consider a more challenging problem setting where we directly regress the absolute root poses of two-hands with extended forearm at high resolution from egocentric view.  
As existing datasets are either infeasible for egocentric viewpoints or lack background variations, we create a large-scale synthetic dataset with diverse scenarios and collect a real dataset from multi-calibrated camera setup to verify our proposed multi-view image feature fusion strategy.  
To make the reconstruction physically plausible, we propose two strategies: 
(i) a coarse-to-fine spectral graph convolution decoder to smoothen the meshes during upsampling and (ii) an optimisation-based refinement stage at inference to prevent self-penetrations.
Through extensive quantitative and qualitative evaluations, we show that our framework is able to produce realistic two-hand reconstructions and demonstrate the generalisation of synthetic-trained models to real data, as well as real-time AR/VR applications.
\end{abstract}

%% file: tex/intro.tex
\section{Introduction}
$3$D hand pose estimation is a fundamental problem in various downstream applications including augmented and virtual reality (AR/VR) \cite{han2020megatrack,wang2020rgb2hands,mueller2019real,han2022umetrack,tse2019no,feng2023mutual,feng2023diffpose}. Research efforts in designing $3$D hand or hand-object reconstruction networks have shown great effectiveness for single-hand pose estimation \cite{hasson2019learning,ge20193d,tse2022collaborative,tse2022s,doosti2020hope, zheng2022tp}. However, two-hand pose estimation has received relatively little attention. In this paper, as illustrated in \fig{teaser}, we focus on high fidelity two-hand reconstruction from multi-view RGB images. 

\input{fig/teaser/item.tex}

With the availability of the InterHand$2.6$M dataset \cite{moon2020interhand2}, various methods have been recently proposed. More recent methods focused on alleviating the self-similarity problem between interacting hands by exploiting hand part segmentation probability \cite{fan2021learning}, joint visibility \cite{kim2021end}, cascaded refinement modules \cite{zhang2021interacting} or keypoints using Transformer \cite{hampali2022keypoint}. 
Although they perform well on complex configurations, their dependence on root joint alignment does not provide absolute root pose recovery when applied to multi-view scenarios, which is especially crucial for interactions in VR.

Moreover, many immersive AR/VR applications require accurate estimation of two hands including extended forearms as it allows for a more realistic and accurate representation of hand movements and gestures. By including the forearm, the orientation and movement of the hand in relation to the arm can provide important contextual information for the user's actions in the virtual environment. Additionally, it can reduce errors and improve the stability of the tracking system, which is important for maintaining immersion and avoiding disorientation in AR/VR applications. However, as in many other areas of computer vision, there is currently no suitable dataset for a typically supervised deep learning setup. Specifically, existing datasets are either infeasible for egocentric views \cite{garcia2018first} or lack variations in background and lighting conditions \cite{moon2020interhand2}.

In this paper, we propose a spectral graph-based transformer architecture that can reconstruct high resolution two-hand meshes with extended forearms from multi-view RGB images. However, directly extending state-of-the-art methods \cite{lin2021end, lin2021mesh} from single-view to multi-view setting is non-trivial as they consume a substantial number of parameters and are computationally expensive. As pointed out by Cho \etal~\cite{cho2022cross}, the current encoder-based transformers \cite{lin2021end, lin2021mesh} have overlooked the importance of efficient token design. In order to retain spatial information in image features and avoid passing redundant global image features to the transformer encoder, we propose a soft-attention-based multi-view image feature fusion strategy to obtain region-specific features. Further, as a hand mesh in essence is graph-structured data, we incorporate properties of graph Laplacian from spectral graph theory \cite{chung1997spectral} into the design of our transformer-based network architecture. To obtain realistic hand reconstructions, we leverage a hierarchical graph decoder and propose an optimisation-based refinement stage during inference to prevent self-penetrations. We show that each technical component above contributes meaningfully in our ablation study.

Besides the expensive computational cost, an additional challenge is the lack of high fidelity $3$D mesh ground truth for training such a model. As manually annotating $3$D hand meshes on real data is extremely laborious and expensive, we create a large-scale synthetic multi-view dataset containing realistic hand motions from egocentric camera viewpoints rendered under a large variation of background and illumination.
To generalise and evaluate on real-world data, we further collected real data from a precisely calibrated multi-view studio with $18$ number of high-resolution cameras. We use an automatic approach for registering meshes and obtain ground truth hand poses. Our proposed method trained on both datasets shows robust performance on challenging scenarios and can serve as a strong baseline.

Our contributions are the following: 
\begin{enumerate}
    \itemsep0em 
    \item We propose a novel end-to-end trainable spectral graph-based transformer for high fidelity two-hand reconstruction from multi-view RGB image. 
    \item We design an efficient soft attention-based multi-view image feature fusion in which the resulting image features are region-specific to segmented hand mesh. We further demonstrate a minimal reduction of $35\%$ in the model size with this approach.
    \item We introduce an optimisation-based method to refine physically-implausible meshes at inference.
    \item We create a large-scale synthetic multi-view dataset with high resolution $3$D hand meshes and collect real dataset to verify our proposed method.
\end{enumerate}

\input{fig/framework/item.tex}

%% file: fig/teaser/item.tex
\begin{figure}
\centering
\includegraphics[width=1\linewidth]{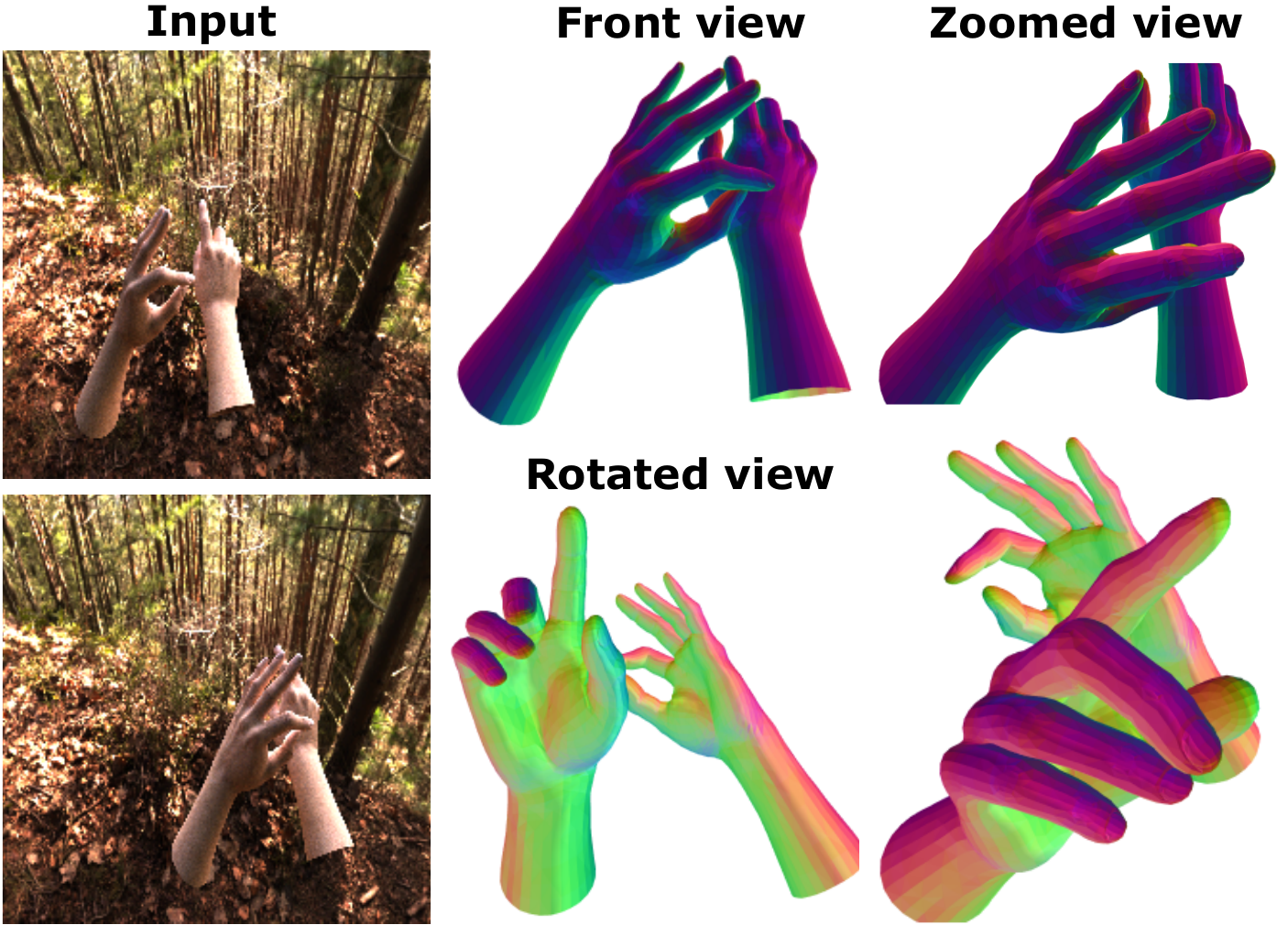}
\caption{Our method jointly reconstructs high fidelity two-hand meshes from multi-view RGB image. The input images shown here are from our synthetic dataset, which contains challenging video sequences of two hands rendered into egocentric views.}
\vspace{-0.4cm}

\label{fig:teaser}
\end{figure}

%% file: fig/framework/item.tex
\begin{figure*}[ht]
\centering
\includegraphics[width=0.95\linewidth]{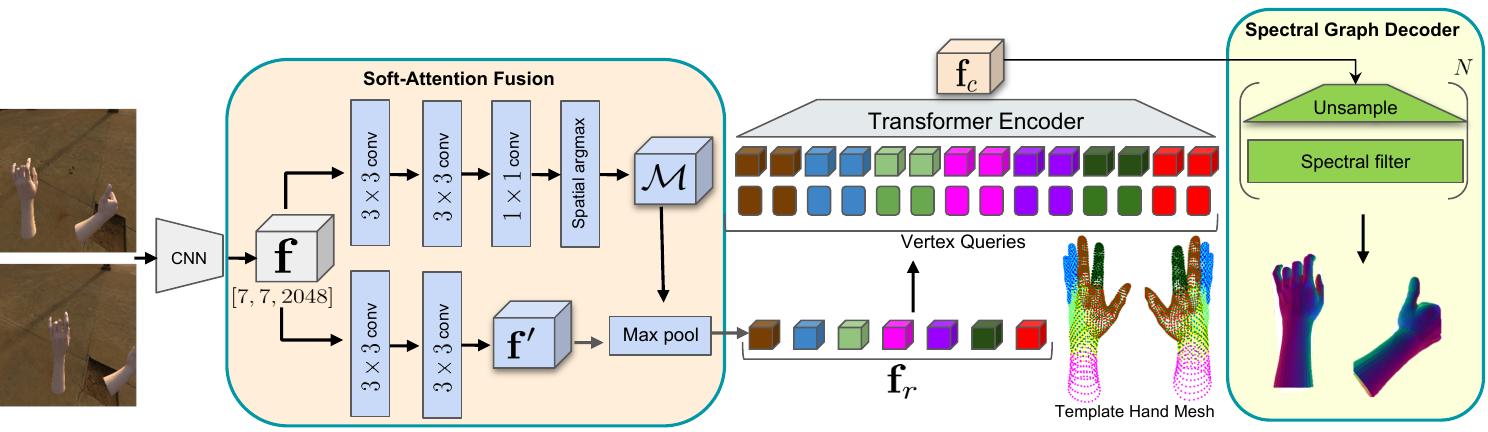}
% \includesvg[width=1\linewidth]{fig/framework/framework.svg}
\vspace{-0.2cm}
\caption{A schematic illustration of our framework. Given multi-view RGB images, we extract volumetric features $\mathbf{f}$ with a shared CNN backbone. The soft-attention fusion block generates the attention mask $\mathcal{M}$ and finer image features $\mathbf{f}'$ through multiple upsampling and convolution blocks. Region-specific features $\mathbf{f}_{r}$ are computed by first aggregating $\mathbf{f}'$ along the feature channel dimension via the attention mask $\mathcal{M}$, followed by a max-pooling operation across multi-view images to focus on useful features. Then, we apply mesh segmentation via spectral clustering on template hand meshes and uniformly subsample them to obtain coarse meshes. We perform position encoding by concatenating coarse template meshes to the corresponding region-specific features $\mathbf{f}_{r}$, \ie matching colored features to mesh segments. Finally, our multi-layer transformer encoder takes the resulting features as input and outputs a coarse mesh representation $\mathbf{f}_{c}$ which is then decoded by a spectral graph decoder to produce the final two-hand meshes at target resolution. Here, each hand contains $4023$ vertices.
}
\vspace{-0.4cm}
\label{fig:framework}
\end{figure*} 

%% file: tex/related.tex
\section{Related Work}
Our work tackles the problem of two-hand reconstruction from multi-view images. We first review the literature on \emph{Hand reconstruction} from RGB images. Then, we focus on the line that leverages \emph{Transformer} on human body/hand reconstruction. Finally, we provide a brief review on existing \emph{Hand datasets}.

\vspace{-0.1cm}
\paragraph{Single-Hand Reconstruction.}
Many approaches have already been proposed for hand pose estimation from either RGB images or depth maps. Here we focus mainly on works that reconstruct full hand meshes. As most earlier methods do not have access to suitable datasets which contain dense $3$D ground truths, the parametric hand model MANO \cite{romero2017embodied} provides an alternative as its input parameters can be directly regressed and it contains prior $3$D information about hand geometry. Hence, discriminative approaches \cite{hasson2019learning,baek2019pushing,boukhayma20193d} which directly predict input hand pose and shape parameters from single images have been a popular approach. \cite{moon2020deephandmesh} learns the pose and shape corrective parameters to reconstruct high-fidelity hand meshes with multi-view depth images. 
Concurrently, non-parametric methods \cite{ge20193d,kulon2020weakly} leveraging graph convolutions demonstrate the ability to directly regress $3$D hand meshes. 
Recently, there is an increasing interest in representing object shapes by learning an implicit function \cite{atzmon2020sal, chen2019learning, mescheder2019occupancy}. By extending NASA \cite{deng2020nasa}, \cite{karunratanakul2021skeleton} is a keypoint-driven implicit hand representation and \cite{corona2022lisa} disentangles shape, color and pose representations.

In contrast to the above mentioned approaches, we propose in this work a non-parametric method that can be extended to multi-view settings and reconstruct high-fidelity hand meshes with extended forearms. Further, our approach is not restricted to a single subject, which is the primary constraint of \cite{moon2020deephandmesh}. Recent work on multi-view fusion leverage epipolar geometry \cite{qiu2019cross,he2020epipolar} to aggregate estimated $2$D heatmaps from different views or Feature Transform Layers (FTL) \cite{remelli2020lightweight, han2022umetrack} to learn camera geometry-aware latent features. However, these methods require known camera parameters and the feature matching process based on feature similarity is sensitive to occlusion. Our work introduces a soft-attention mechanism coupled with transformer tokenisation which is robust to occlusion and aggregates multi-view image features efficiently.

\vspace{-0.1cm}
\paragraph{Two-Hand Pose Estimation.} 
Pose estimation of interacting hands can be broadly classified as discriminative and generative (or hybrid) approaches. Early methods typically follow hybrid approaches by leveraging visual cues detected by discriminative methods, followed by model fitting. \cite{ballan2012motion, tzionas2016capturing} apply collision optimisation terms and physical modelling on detected fingertips. \cite{han2020megatrack, mueller2019real, wang2020rgb2hands, han2022umetrack} extract image features or keypoints from RGB or depth images and fit hand models with physical constraints. Recent large-scale datasets of interacting hands \cite{moon2020interhand2} have led to the emergence of fully discriminative methods \cite{kim2021end, li2022interacting, hampali2022keypoint, rong2021monocular, zhang2021interacting} that jointly estimate the $3$D joint locations or hand model parameters from a single RGB image. In this work, we target to reconstruct two hands at high resolution while keeping the model complexity low.

\vspace{-0.1cm}
\paragraph{Transformer in 3D Vision.}
Transformer-based architectures have been gaining increasing popularity in the vision community. Here we focus on methods that reconstruct the human body or hands from RGB images and refer readers to \cite{khan2022transformers} for a detailed survey. In a closely related work, \cite{lin2021end} uses cascaded Transformer encoders to reconstruct the human body and hands from a single RGB image and achieves state-of-the-art performance. \cite{lin2021mesh} extends \cite{lin2021end} with graph convolutions along the Transformer encoder. \cite{cho2022cross} improves the efficiency of \cite{lin2021end} by disentangling image encoding and mesh estimation via an encoder-decoder architecture. \cite{hampali2022keypoint} extends "Detection Transformer" \cite{carion2020end} with hand-object pose estimations. While these works are targeted at a single image and their extension to the multi-view setting is non-trivial due to large number of learnable parameters, our architecture is designed to efficiently reconstruct two hands at high resolution.

\vspace{-0.1cm}
\paragraph{Hand Pose Datasets.} 
The success of discriminative methods depend on the availability and variability of hands interaction datasets. \cite{moon2020interhand2} introduces the large-scale dataset InterHand$2.6$M with closely interacting hand motions using a semi-automatic annotation process with multiple high-resolution RGB cameras. However, it contains minimal background and lighting variation. FreiHAND \cite{zimmermann2019freihand} includes wider background variation but the dataset is limited to a single hand and third-person views. Therefore, the above datasets are not suitable for the egocentric two-hand reconstruction task. The available egocentric datasets are limited by either visible markers \cite{garcia2018first} or constrained lab environments \cite{kwon2021h2o}. 
This motivates us to create a large-scale egocentric synthetic dataset with improved environment and lighting variations. 
To validate our proposed multi-view fusion strategy, we further collect a real dataset with more challenging camera viewing angles. Altogether, both synthetic and real datasets contain diverse egocentric multi-view and -frame data points with multiple subjects. Our architecture demonstrates promising performance on both datasets and constitutes a strong baseline.

%% file: tex/method.tex
\section{Method}

As shown in \fig{framework} 
, our architecture first passes $N$ number of multi-view RGB input images $\bx \in \mathbb{R}^{N\times224\times224\times3}$ 
to a shared CNN backbone to extract volumetric features $\mathbf{f}\in\mathbb{R}^{N\times7\times7\times2048}$, \ie features before the global average pooling layer for ResNet
\cite{he2016deep}. The volumetric features are then fed into a soft attention-based multi-view feature encoder and output $K$ region-specific features $\mathbf{f}_{r} \in \mathbb{R}^{ K\times C}$ where $C$ refers to the feature channel size. The Transformer encoder takes $\mathbf{f}_{r}$ together with template hand meshes $\bm' \in \mathbb{R}^{V'\times3}$ and outputs a coarse mesh representation $\mathbf{f}_{c}\in\mathbb{R}^{V'\times F}$. Finally, the spectral graph decoder generates hand meshes $\bm \in \mathbb{R}^{V\times3}$ by upsampling on $\mathbf{f}_{c}$ where $V \gg V'$. With slight abuse of notation, $\mathbf{m}$ can either be two-hand or single-hand depending on the application.

In the following, we detail the multi-view image feature encoder in Section \ref{sec:encoder}, the spectral graph convolution decoder in Section \ref{sec:decoder} and the loss used for training in Section \ref{sec:training}. We also present an optimisation-based refinement procedure at inference time in Section \ref{sec:inference}.

\subsection{Notations} 
A $3$D mesh can be represented as an undirected graph $\mathcal{G} = (\mathcal{V}, \mathcal{E}, \mathbf{A})$, where $\mathcal{V}$ is a node set and $\mathcal{E}$ is an edge set.
An adjacency matrix $\mathbf{A} \in \mathbb{R}^{|\mathcal{V}| \times |\mathcal{V}|}$ encodes information of pairwise relations between nodes.
A degree matrix $\mathbf{D} \in \mathbb{R}^{|\mathcal{V}| \times |\mathcal{V}|}$ is a diagonal matrix whose diagonal element $\mathbf{D}_{ii} = \sum_{j} \mathbf{A}_{ij}$ refers to the degree value of each node.
An essential operator in spectral graph theory \cite{chung1997spectral} is the graph Laplacian $\mathbf{L}$, whose definition is $\mathbf{L} = \mathbf{D} - \mathbf{A}$, and $\mathbf{L} =\mathbf{U}\mathbf{\Lambda}\mathbf{U}^T$ where the graph Laplacian can be diagonalised by the Fourier basis $\mathbf{U} = [\mathbf{u}_1,\ldots,\mathbf{u}_{|\mathcal{V}|}] \in \mathbb{R}^{|\mathcal{V}| \times |\mathcal{V}|}$ and $\mathbf{\Lambda} = diag([\lambda_1,\ldots,\lambda_{|\mathcal{V}|})] \in \mathbb{R}^{|\mathcal{V}| \times |\mathcal{V}|}$ where $\{\mathbf{u}_i\}_{i=1}^{|\mathcal{V}|}$ are the eigenvectors and 
$\{\lambda_i\}_{i=1}^{|\mathcal{V}|}$ are the non-negative eigenvalues of graph Laplacian ($0 = \lambda_1 \leq \ldots \leq \lambda_{|\mathcal{V}|}$).

\subsection{Multi-View Image Feature Encoder}\label{sec:encoder} 
To extend transformer-based architecture to multi-view settings and prevent concatenating global image features in an overly-duplicated way as in \cite{lin2021end}, we propose a simple \emph{soft-attention fusion} strategy to better aggregate features across multiple views and attend to different hand parts via \emph{mesh segmentation}. The resulting $K$ region-specific features $\mathbf{f}_{r}$ defined by \emph{spectral clustering} are fed to our \emph{Transformer encoder} to obtain a coarse mesh representation $\mathbf{f}_{c}$.

\paragraph{Soft-Attention Fusion.} 
Given volumetric features $\mathbf{f}$, we do not apply any pooling operations to avoid losing spatial information as shown by our experiments (see Table \ref{table:abl_multi_view}).
Instead, we propose to aggregate multi-view features with a soft-attention mask. We first obtain a finer representation of $\mathbf{f}$, denoted as $\mathbf{f}' \in \mathbb{R}^{N\times(H\times W)\times C}$, by feeding it through two blocks each comprised of $2$D upsampling with bilinear interpolation, $3\times3$ convolution layers, batch-normalisation \cite{ioffe2015batch} and ReLU.
The soft-attention mask $\mathcal{M} \in \mathbb{R}^{N\times(H\times W)\times K}$ is obtained by applying 1) $K$ $1\times1$ convolutional filters to reduce the feature channel of $\mathbf{f}'$ to $K$ and 2) spatial soft arg-max which determines the image-space point of maximal activation in each $C$. At this stage, we can compute per-frame features $\mathbf{f}'' \in \mathbb{R}^{N\times K\times C}$ by $\mathbf{f}''=\mathcal{M}^{\text{T}}\mathbf{f}'$. We finally obtain region-specific features $\mathbf{f}_{r} \in \mathbb{R}^{K\times C}$ by max-pooling along the second dimension of $\mathbf{f}''$. There are two intuitions to this design: 1) the corresponding feature with higher attention weight contributes more to the final feature representation spatially and 2) max-pooling allows efficient feature selection across multiple views and it does not overfit to any multi-camera configurations as CNN filters are shared across all views.

\paragraph{Mesh Segmentation via Spectral Clustering.} 
Inspired by \cite{liu2004segmentation}, we perform spectral clustering to obtain a 3D mesh segmentation. Different from \cite{liu2004segmentation}, we apply the eigen-decomposition of the graph Laplacian $\mathbf{L}$ instead of the affinity matrix (which encodes pairwise point affinities with exponential kernel), followed by $k$-means clustering into $K$ clusters. 
With this approach, one can efficiently segment any template model mesh without manual effort. In addition, as shown in \fig{cluster}, the segmented mesh does not exhibit clear-cut boundaries, and certain clusters are scattered throughout the entire surface. This is in spirit similar to mask vertex modeling occlusions in \cite{lin2021end} by encouraging the transformer to consider other relevant vertex queries.

\paragraph{Transformer Encoder.} 
We now concatenate the $C$-dimensional region-specific image features to the corresponding $K$ clusters of the segmented template hand mesh. The resulting features $\mathbf{f}_{t} \in \mathbb{R}^{V'\times (C+3)}$ 
\footnote{$+3$ refers to the 3D positions of mesh template.} 
are fed into a multi-layer transformer encoder with progressive dimensionality reduction as described in \cite{lin2021end}. The output is a coarse mesh representation $\mathbf{f}_{c}\in\mathbb{R}^{V'\times F}$, where $F = (C+3)/2^n$ with $n$ layers of transformer encoder. 

\paragraph{Theoretical Motivations.} 
Since there do not exist explicit coordinate systems as in grid graphs, aligning nodes of highly irregular graphs in nature is a non-trivial problem.
Recent studies \cite{dwivedi2020benchmarking, dwivedi2021generalization, kreuzer2021rethinking} attempted to encode positional information by leveraging the spectral domain in a way that nearby nodes have similar values and distant nodes have different values.
This can be achieved as the eigenvectors of the graph Laplacian can be interpreted as the generalised concepts of sinusoidal functions of positional encoding in Transformers \cite{vaswani2017attention}. 
For instance with any given graph, nodes close to each other can be assigned values of similar positional features as the smaller the eigenvalue of the graph Laplacian $\lambda_i$ (closer to $0$) the smoother the coordinates of the corresponding eigenvector $\mathbf{u}_i$. 

\input{fig/clustering/item}

\subsection{Spectral Graph Decoder}\label{sec:decoder}
We find that simply relying on fully-connected layers to upsample meshes to target resolution is insufficient as this process introduces instability and disruption to the mesh, even with mesh regularisation loss terms \cite{hasson2019learning} (see \fig{decoder}). Therefore, we propose to couple the fully-connected layers for upsampling with \emph{spectral filtering} as meshes can be treated as graph signals $\mathbf{f}_{c} = (f_{1}, \cdots, f_{V'}) \in\mathbb{R}^{V'\times F}$, \ie $V'$ vertices with $F$-dimensional features for batch size of $1$.

\paragraph{Spectral Filtering.}
As spectral graph convolution can be defined via point-wise products in the transformed Fourier space, graph signals $\mathbf{f}_{c}$ filtered by $g_{\theta}$ can then be expressed as $\mathbf{U}g_{\theta}(\mathbf{\Lambda})\mathbf{U}^T\mathbf{f}_{c}$ where $g_{\theta}(\mathbf{\Lambda})$ is a diagonal matrix with Fourier coefficients. To ensure that the spectral filter corresponds to a meaningful convolution on the graph, a natural solution is to parameterise based on the eigenvalues of the Laplacian. Here we have used the Chebyshev polynomial parametrisation of $g_{\theta}(\mathbf{L})$ for fast computation and readers are referred to \cite{defferrard2016convolutional} for more details.

\paragraph{Architecture.}
Similar to \cite{ge20193d, choi2020pose2mesh}, our decoder is based on a hierarchical architecture where the mesh is recovered by using fully-connected and graph convolution layers for upsampling. We followed \cite{ge20193d, defferrard2016convolutional} to pre-computed coarse graphs using \cite{dhillon2007weighted} and used the third-order polynomial in the Laplacian. Given the coarse mesh representation $\mathbf{f}_{c}\in\mathbb{R}^{V'\times F}$, the key idea here is to smooth out the $F$-dimensional values after upsampling on $V'$.
Note that the fundamental difference between this approach and the one in \cite{ge20193d, choi2020pose2mesh} is that they have a hierarchical architecture on both $V'$ and $F$ dimensions, we apply only spectral filtering and keep $F=3$ constant which massively reduces the model size while maintaining performance (see Table \ref{table:main_syn}).

\paragraph{Discussions.} 
Recall that in Section \ref{sec:encoder}, we leverage the properties of Laplacian eigenvectors to perform spectral clustering. Here we can interpret from a signal processing perspective where the Laplacian eigenvectors define signals that vary smoothly across the graph, with the smoothest signals indicating the coarse community structure of the mesh. 

\subsection{Training}\label{sec:training}
Following \cite{lin2021end,lin2021mesh}, our model can be trained end-to-end with L$1$ losses on $3$D mesh vertices and $2$D re-projection using the predicted camera parameters to improve image-mesh alignment. In addition, we apply an edge length regularisation $\loss_{edge}$ \cite{hasson2019learning} to encourage smoothness of the mesh: 
\vspace{-0.3cm}
\begin{align} 
    \loss_{edge}(\bm) = \frac{1}{\lvert\mathcal{E}_{L}\rvert}\sum_{l\in\mathcal{E}_{L}}\lvert l^2 - \mu(\mathcal{E}_{L}^2)\rvert,
\end{align}
where $\mathcal{E}_{L}$ refers to the set of edge lengths, defined as the L$2$ norms of all edges and $\mu(\mathcal{E}_{L}^2)$ is the average of the squared edge lengths.

\subsection{Mesh Refinement at Inference}\label{sec:inference}
There are two main approaches in the literature for producing realistic mesh reconstruction: learning-based and optimisation-based methods.
Learning-based methods \cite{hasson2019learning, moon2020deephandmesh} approach the problem by proposing various repulsive losses that penalise penetration during training. However, their generalisation ability to other meshes is limited due to the need for mesh-specific pre-computation. In particular, they require manual selection of areas of interest (\ie fingertips and palm) which does not prevent other forms of self-penetrations, such as finger-finger. More importantly, the presence of interpenetration at test time shows that the model is not able to learn the physical rule implicitly \cite{hasson2019learning, tse2022collaborative}. On the other hand, recent optimisation-based methods \cite{grady2021contactopt, tse2022s} leverage contact maps to refine meshes at inference. However, they rely heavily on accurate contact map estimations and are sensitive to initialisation as contact optimisation is local. 

\paragraph{Proposed Method.} To avoid penetrations, we extend the repulsion loss from \cite{hasson2019learning} into an optimisation-based strategy which does not require any form of pre-computation and is more generalisable to other meshes. To identify hand vertices that contribute to collision, we follow \cite{moller1997fast, hasson2019learning} and cast rays from each vertex and count the number of surface intersections. If the number is odd, it indicates penetration. After obtaining the collision mask $\mathcal{M}_{C}$, we compute the nearest point in the source mesh $\bm$ with respect to the set of collision points. If the point and its nearest corresponding point have a different normal, we compute their distance as loss. We minimise this collision loss $\loss_{collision}$ with as-rigid-as-possible (ARAP) \cite{sorkine2007rigid} regularisation on the mesh shape:
% \vspace{-0.1cm}
\begin{align} 
    \loss_{collision}(\bm, \mathcal{M}_{C}) = \sum_{v\in\mathcal{V}} \mathcal{M}_{C}\cdot d(v, \mathcal{V}),
\end{align}
where $\mathcal{M}_{C} = \mathds{1}_{v\in\mathrm{Int}(\mathcal{V})}$ indicates which vertices belong to the interior of the mesh and $d(v, \mathcal{V}) = \min_{v'\in \mathcal{V}} \Vert v - v' \Vert_{2}$
denotes distances from point $v'$ to set $\mathcal{V}$. We show that our method is able to remove self-penetration in \fig{collision} and \ref{fig:mesh_refinement_full}.

\input{fig/qualitative/decoder/item}
\input{fig/qualitative/collision/item}

%% file: fig/clustering/item.tex
\begin{figure}
\centering
% \includesvg[width=0.8\linewidth]{fig/clustering/cluster_1.svg}
\includegraphics[width=0.8\linewidth]{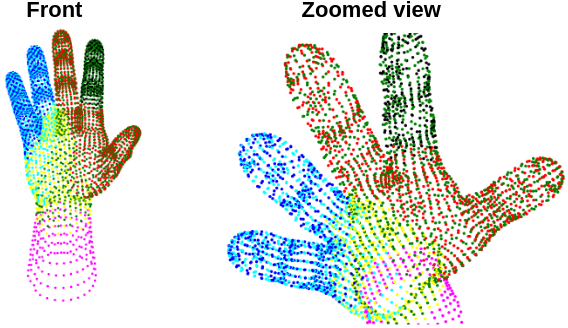}
\caption{
Illustration of mesh segmentation via spectral clustering. Here, we have chosen $K=7$ clusters and each has its own color for a single right-handed template mesh. As shown, there are no rigid boundaries across the mesh.
To prepare inputs $\mathbf{f}_{t}$ to transformer, we decrease the size of this mesh (similarly for left hand) uniformly by a factor of $10$ through subsampling, \ie $V'=V/10$ and concatenate with the corresponding region-specific features $\mathbf{f}_{r}$.
}
\vspace{-0.4cm}
\label{fig:cluster}
\end{figure}

%% file: fig/qualitative/decoder/item.tex
\begin{figure}
\centering
\includegraphics[width=0.9\linewidth]{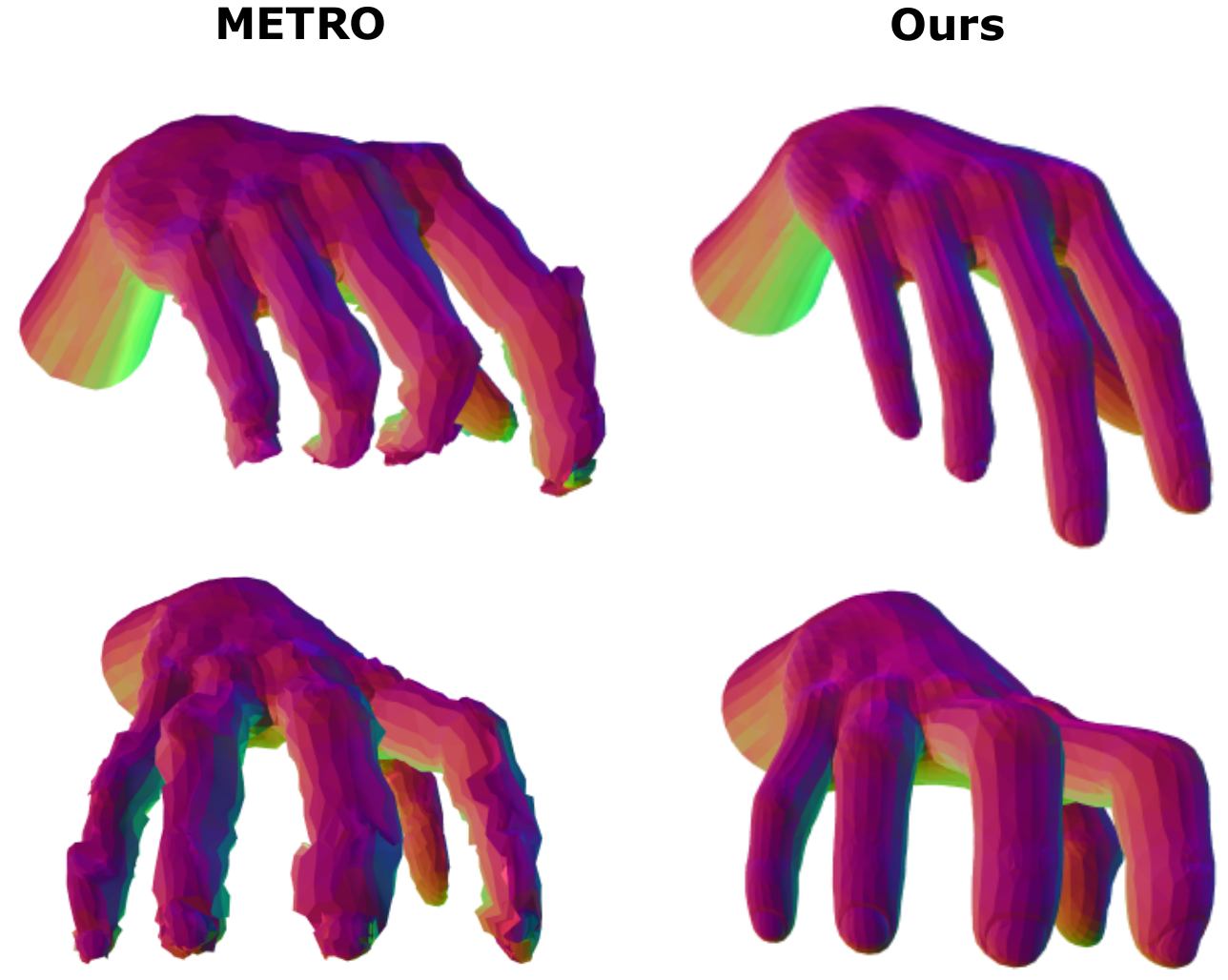}
\caption{Qualitative comparison with METRO \cite{lin2021end}. We show that relying on fully-connected layers to upsample meshes is inadequate for high-resolution mesh reconstruction. In contrast, our spectral graph decoder can accurately capture intricate surface features like nails.}

\label{fig:decoder}
\end{figure}    

%% file: fig/qualitative/collision/item.tex
\begin{figure}
\centering
\includegraphics[width=0.9\linewidth]{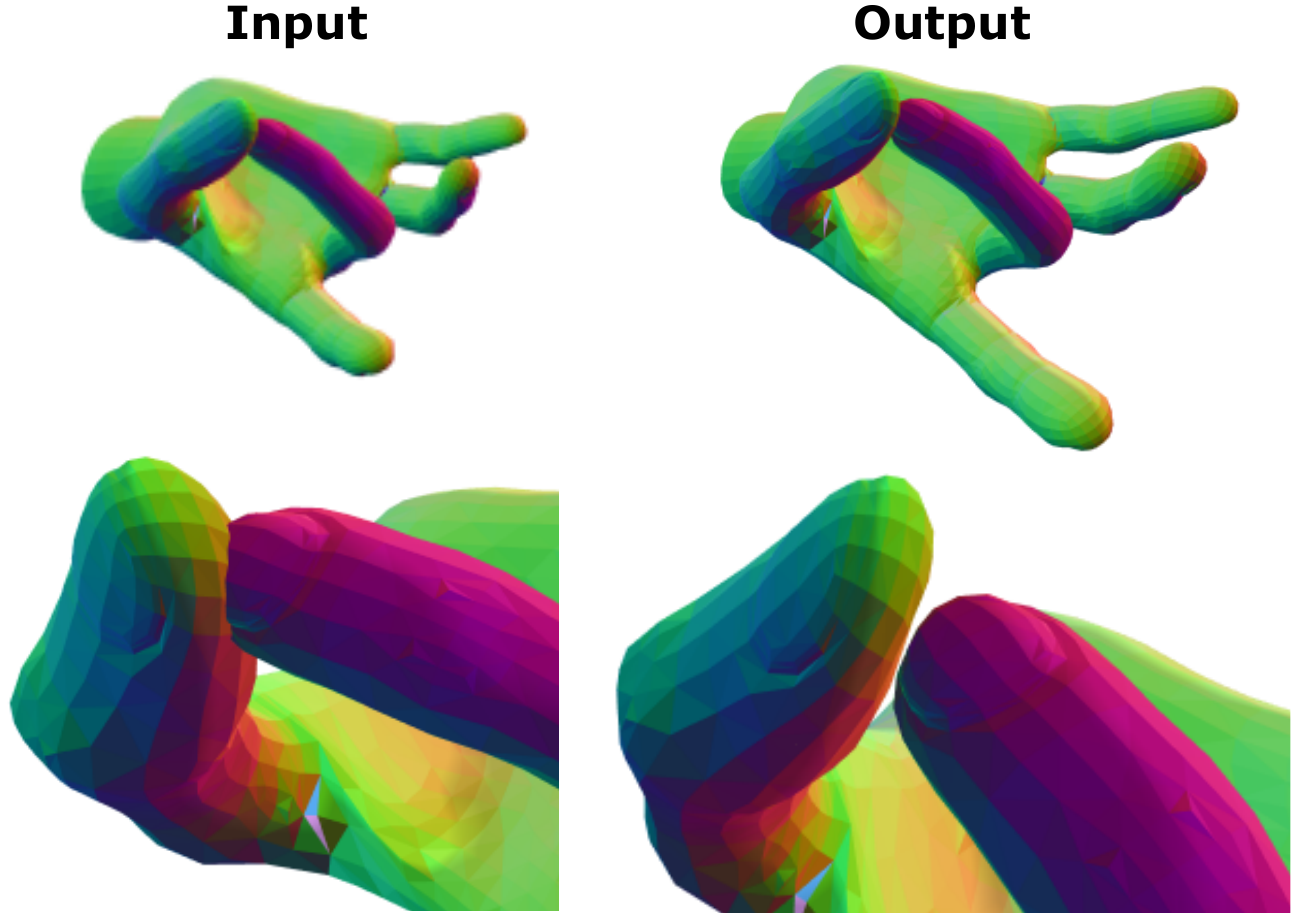}
\caption{Qualitative example of mesh refinement at inference. More examples are available in \fig{mesh_refinement_full}.}
\label{fig:collision}
\vspace{-0.4cm}
\end{figure}    
 

%% file: tex/result.tex
\section{Experiments}
\paragraph{Implementation Details.} We implement our method in TensorFlow. We train all parts of the network simultaneously with the Adam optimiser \cite{kingma2014adam} using a learning rate of $10^{-4}$ when training on the synthetic dataset and $10^{-5}$ when fine-tuning on the real dataset. We use ResNet \cite{he2016deep} pre-trained on ImageNet \cite{russakovsky2015imagenet} for our CNN backbone. For computing region-specific features $\mathbf{f}_{r}$, we set $K=7$ and $C=256$.

\paragraph{Synthetic Dataset Creation.} A large-scale multi-view egocentric dataset with challenging interacting hand motion is required to train our pipeline. However, existing datasets are either infeasible for egocentric views or lack variations in terms of background and lighting conditions, which are crucial for AR/VR applications. In addition, the ground-truth mesh in \cite{moon2020interhand2} contains $5mm$ fitting error which is sufficient to cause inaccuracy in hand tracking or misalignment when interacting with virtual objects.
Therefore, it is not suitable to validate our approach. To this end, we create a new large-scale synthetic egocentric dataset with two high-fidelity $3$D hand meshes. In particular, we purchase commercial $3$D hand models for both left and right hands. Each of them contains $4023$ vertices and $4008$ quad faces. By dividing each quad into two triangles, they can be decomposed into $8016$ triangular faces. The size and textures of the hand models can vary, and it can also be rigged with $3$D joint locations. We apply photorealistic textures as well as natural lighting using High-Dynamic-Range (HDR) images. To create realistic hand motions, we generate $100$ poses and interpolate between pairs of these poses randomly over $1000$ frame sequences, \ie pose pair is swapped every $10$ frames. Then, we render hands onto $480$ $4$K backgrounds using the Cycles renderer \cite{cycles}. Our synthetic dataset comprises $1$M data points. Each data point consists of 3D annotations for two hands, rendered in two egocentric views to simulate a customised camera headset scenario. We divide the entire dataset into $4$ groups and use $25\%$ randomly for testing and provide additional image examples in \fig{synthetic_examples}.

\input{fig/qualitative/synthetic_examples/item}

\paragraph{Real Dataset Collection.}
Our real hand data is captured from a multi-view stereo system with $18$ synchronised Z-Cams. We build a NeRF-based~\cite{barron2022mip,muller2022instant} reconstruction pipeline that simultaneously reconstructs and disentangles the foreground and the background, from which a meshing module extracts hand meshes as ground-truth target mesh. We then apply a mesh and tetrahedral registration approach~\cite{smith2020constraining} that registers a template hand model~\cite{xu2020ghum} to get a well-registered mesh for each reconstruction.

\paragraph{Baselines.} 
We compare with the state-of-the-art transformer-based architecture METRO \cite{lin2021end} that is explicitly designed for human body mesh reconstruction. We follow their official implementation and use $3$ transformer encoder layers. We attempted to include other strong baselines to compare with. However, given our challenging multi-view settings on reconstructing high resolution two-hand meshes, existing methods are constrained by either 1) relying on intermediate $3$D pose supervision \cite{moon2020i2l, choi2020pose2mesh} or 2) directly regressing a low resolution parametric hand model \cite{hasson2019learning,zimmermann2019freihand, baek2019pushing, boukhayma20193d, tse2022collaborative}. In addition, the direct extension of METRO \cite{lin2021end}, Mesh Graphormer \cite{lin2021mesh}, is infeasible to extend to multi-view settings as they tokenised volumetric features which increases the computational complexity of each transformer layer quadratically \cite{cho2022cross}. Therefore, by considering that the performance gain is minor, we pick METRO \cite{lin2021end} as a strong baseline and study in-depth. We provide more details about the baselines in the supplementary materials. 
 
\paragraph{Evaluation Metric.} We report the Mean-Per-Vertex-Error (MPVE) in $mm$ to evaluate the \emph{hand reconstruction error}. MPVE measures the mean Euclidean distances between the ground-truth vertices and the predicted vertices.

\paragraph{Quantitative Comparison.} 
We perform a quantitative comparison for the \emph{two-hand reconstruction} task on our synthetic dataset. As previously mentioned, comparing with other methods is not straightforward since the majority of existing methods focus on single-view settings. We extend METRO \cite{lin2021end} to the multi-view setting by applying max-pool to image features and concatenating them with vertex queries. For fair comparisons, we share the same hyperparameter setting for the multi-layer transformer encoder and report the results in Table \ref{table:main_syn}. In this experiment, we downsample the template hand mesh by $10$ times. To provide more context regarding the difficulty of our synthetic dataset, we also experiment on a parametric baseline. We create a parametric hand model which uses the $200$-dimensional PCA (principal component analysis) subspace from $25\%$ of the training data. We use ResNet-$50$ as the backbone for all models in Table \ref{table:main_syn} and the parametric baseline has an MLP head to predict the input parameters to recover the hand mesh. Our method significantly outperforms both baseline methods with less than half the model size of METRO. We show that our soft-attention feature fusion strategy coupled with mesh segmentation using spectral clustering can effectively reduce the feature channel size from $2048$ to $256$ without performance drop. In addition, the performance of the parametric baseline is in-line with existing single-hand benchmarks. Lastly, we report results on single-view benchmark FreiHAND \cite{zimmermann2019freihand} in Table \ref{table:benchmarks}.

\input{tbl/main_synthetic}
\input{tbl/rebuttal/single_view/item}

\paragraph{Ablation Study.} To motivate our design choices, we present a quantitative evaluation of our method with various components disabled. We validate that each of our proposed technical component contributes meaningfully. 

\emph{Effects of Multi-View Feature Fusion.} 
Table \ref{table:abl_multi_view} shows the results of varying number of spectral clusters $K$ (full results in supp.). The combination of soft-attention fusion and mesh segmentation consistently improves the performance. As our synthetic dataset contains only two views, we further verify our multi-view fusion strategy on more challenging viewing angles captured in our real dataset. We also demonstrate that our method does not overfit to camera setup when tested on unseen camera views in Table \ref{table:abl_handbooth}. In these experiments, we divide the $18$ camera views into $2$ groups, \ie the first group contains the first $15$ camera views and the second group contains the last $3$ unseen camera views for evaluation. Note that the evaluation group contains the only egocentric view. We consider $3$ experimental settings by varying number of camera views present in the training data: (a) $15$ views, (b) $6$ views and (c) $3$ views in Table \ref{table:abl_handbooth}.

\input{tbl/abl_multi_view}
\input{tbl/abl_handbooth}
\input{tbl/abl_graph_decoder}

\input{fig/qualitative/challenging/item}
\input{tbl/smaller_model}
\input{fig/qualitative/handbooth/item}

\emph{Effects of Spectral Filters.} We experiment with three commonly-used spectral filters: Gaussian $g_{gau}$, Laplacian $g_{lap}$ and Chebyshev filters $g_{cheb}$ with varying order of polynomials in Table \ref{table:abl_graph_decoder}. We provide implementation details and additional qualitative examples in the supplementary. We choose $g_{cheb}$ for efficiency and do not find increasing order of polynomials improves performance further.

\vspace{-0.1cm}
\paragraph{Synthetic to Real Transfer.}
Large-scale synthetic dataset can be used to pre-train models in the absence of suitable real datasets. 
In the following, we demonstrate a use case of deploying our pre-trained model which was trained on synthetic data to general real images. First, we train on a subset of our real dataset and freeze all part of the pre-trained model except CNN backbone. As the registered meshes for our real dataset have a different mesh topology, we drop the pink cluster of our original template mesh (shown in \fig{cluster}) before training. This approach is analogous to MANO-based $3$D hand mesh estimation methods \cite{baek2019pushing, boukhayma20193d} and has been demonstrated in \cite{moon2020deephandmesh}. We show that our model can generalise to in-the-wild-images in \fig{handbooth}.

\vspace{-0.05cm}

\paragraph{Model Compression.} 
We are interested in finding the minimal size to which model can be compressed while maintaining the METRO baseline performance in Table \ref{table:main_syn}. First, we perform computational complexity analysis on the self-attention layer in the Transformer encoder. There are two key steps to compute self-attention: 1) linear projection to $C$-dimensional query, key and value matrices from $V'$ vertex queries requires $O(V'C^2)$ and softmax operation for layer output requires another $O(V'^2C)$. The total computational complexity of each transformer layer is therefore quadratic no matter whether $V'$ or $C$ dominates. Recall that as we uniformly subsample template hand meshes by $10$ times to obtain $V'=804$ and $C$ is empirically set to $256$ in earlier sections, $V'$ contributes more to overall complexity as it is dominantly larger. In the following experiments, we gradually decrease $V'$ before $C$ while using a variant of EfficientNet \cite{tan2019efficientnet}. We present main results in Table \ref{table:smaller_model}, while full results are provided in the supplementary.

%% file: fig/qualitative/synthetic_examples/item.tex
\begin{figure}
\centering
\includegraphics[width=1\linewidth]{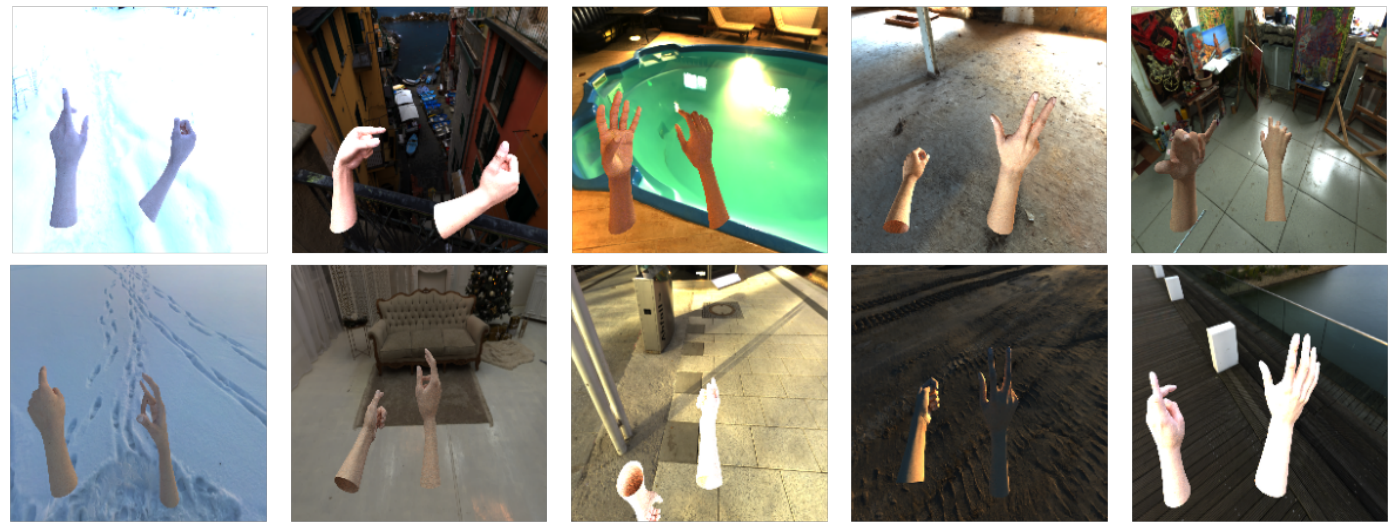}
\caption{Qualitative examples of our synthetic dataset. The scenes are rendered with large variations in lighting, texture and background.}
\vspace{-0.6cm}
\label{fig:synthetic_examples}
\end{figure}  

%% file: tbl/main_synthetic.tex
\newcolumntype{C}{>{\centering\arraybackslash}X}
\begin{table}
\begin{center}
\caption{Error rates on our synthetic dataset. Parametric baseline is trained and tested on right hand only.
Our proposed method achieves strong performance with less than half the METRO model size.
}
\label{table:main_syn}
% \vspace{-0.2cm}
\resizebox{0.9\linewidth}{!}{%
\begin{tabularx}{\linewidth}{l | C | C }
\toprule
 & Hand error & \# Params \\ 
\midrule 
Parametric baseline & 24.5 & 40.9M\\
METRO \cite{lin2021end} & 7.09 & 116.1M \\
\midrule
Ours (w/o graph decoder) & 3.72 & 75.2M\\
Ours  & \textbf{1.38} & 58.3M \\
\bottomrule
\end{tabularx}
}
\end{center}
\vspace{-0.4cm}
\end{table}

%% file: tbl/rebuttal/single_view/item.tex
\newcolumntype{C}{>{\centering\arraybackslash}X}
\begin{table}[t]
\begin{center}
% \vspace{-0.3cm}
\caption{Error rates (in $mm$) on FreiHAND.}
\label{table:benchmarks}
% \vspace{-0.3cm}
\resizebox{0.85\linewidth}{!}{%
\begin{tabularx}{\linewidth}{r | r| r| r}
\toprule
 & METRO \cite{lin2021end} & Graphormer \cite{lin2021mesh} & Ours  \\ 
\midrule 
PA-MPVPE $\downarrow$ & 6.7 & 5.9 & \textbf{5.5}\\
PA-MPJPE $\downarrow$ & 6.8 & 6.0 & \textbf{5.6}\\
\bottomrule
\end{tabularx}%
}
\end{center}
\vspace{-0.6cm}
\end{table}

%% file: tbl/abl_multi_view.tex
\newcolumntype{C}{>{\centering\arraybackslash}X}
\begin{table}
\begin{center}
\caption{Performance of different multi-view fusion strategies. We report hand error for both settings. $K$ refers to the number of clusters for template hand mesh. Note that we do not include spectral filtering in the graph decoder here.
}
\label{table:abl_multi_view}
% \vspace{-0.2cm}
\resizebox{0.9\linewidth}{!}{%
\begin{tabularx}{\linewidth}{l | C | C }
\toprule
 & Single-view & Multi-view \\ 
\midrule 
METRO \cite{lin2021end} & 10.87 & - \\
METRO \cite{lin2021end} + avg. pool & - & 8.71 \\ 
METRO \cite{lin2021end} + max pool & - & 7.09 \\ 
\midrule
Ours ($K=1$) & - & 6.59\\
% Ours ($K=2$) & - & 5.71\\
% Ours ($K=3$) & - & 5.19\\
Ours ($K=4$) & - & 4.79\\
% Ours ($K=5$) & - & 4.59\\
% Ours ($K=6$) & - & 5.28\\
Ours ($K=7$) & - & \textbf{3.72} \\
% Ours ($K=8$) & - & 3.79\\
\bottomrule
\end{tabularx}
}
\end{center}
\vspace{-0.4cm}
\end{table}

%% file: tbl/abl_handbooth.tex
\newcolumntype{C}{>{\centering\arraybackslash}X}
\begin{table}
\begin{center}
\caption{Ablations on multi-view feature fusion. We report hand error on $3$ experimental settings ((a)-(c)) with different feature fusion strategies. We keep the same transformer and spectral graph decoder for all fusion strategies.
}
\label{table:abl_handbooth}
% \vspace{-0.2cm}
\resizebox{0.95\linewidth}{!}{%
\begin{tabularx}{\linewidth}{l | C  C  C}
\toprule
 & (a) & (b) & (c)  \\ 
 \midrule
 \footnotesize{Direct concatenation} & \footnotesize11.3 & \footnotesize13.9 & \footnotesize14.7  \\
\footnotesize{Max pool} & \footnotesize9.3 & \footnotesize11.2 & \footnotesize12.5  \\
\footnotesize{Soft-attention + max pool} & \footnotesize\textbf{6.7} & \footnotesize\textbf{7.2} & \footnotesize\textbf{7.4}  \\
\bottomrule
\end{tabularx}
}
\end{center}
\vspace{-0.4cm}
\end{table} 

%% file: tbl/abl_graph_decoder.tex
\newcolumntype{C}{>{\centering\arraybackslash}X}
\begin{table}
\begin{center}
\caption{Ablations of different spectral filters.  
}
\label{table:abl_graph_decoder}
% \vspace{-0.2cm}
\resizebox{0.95\linewidth}{!}{%
\begin{tabularx}{\linewidth}{l | C  C | C  C  C  C }
\toprule
 & $g_{gau}$ & $g_{lap}$ & $g_{cheb_3}$ & $g_{cheb_4}$ & $g_{cheb_5}$ & $g_{cheb_6}$ \\ 
 \midrule
\footnotesize{Hand error} & \footnotesize1.56 & \footnotesize1.45 & \footnotesize1.38 & \footnotesize1.47 & \footnotesize1.38 & \footnotesize1.39 \\
\bottomrule
\end{tabularx}
}
\end{center}
\vspace{-0.4cm}
\end{table} 

%% file: fig/qualitative/challenging/item.tex
\begin{figure*}[ht]
\centering
\includegraphics[width=1\linewidth]{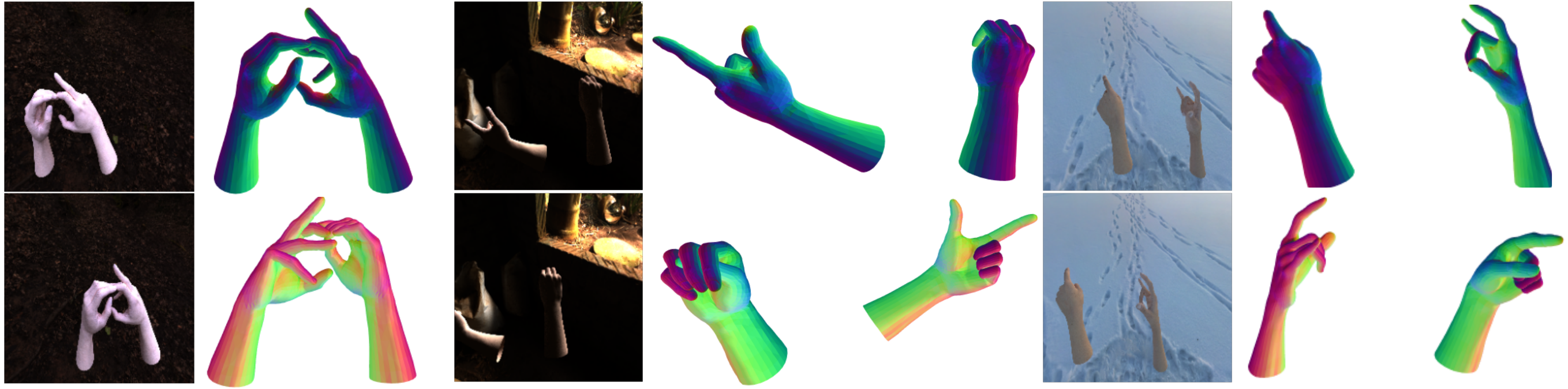}
\vspace{-0.4cm}
\caption{
Qualitative examples on our synthetic dataset. Our model shows robustness to various extreme self-occlusions (left), lighting conditions (middle) and scenes (right). Fine mesh reconstruction details are well-preserved in all scenarios.
}
\vspace{-0.4cm}
\label{fig:challenging}
\end{figure*}

%% file: tbl/smaller_model.tex
\newcolumntype{C}{>{\centering\arraybackslash}X}
\begin{table}
\begin{center}
\caption{Ablations of different backbones and hyperparameters. We denote $P_{cnn}$ and $P_{total}$ to be the number of parameters for CNN backbone and total model, respectively.
}
\label{table:smaller_model}
% \vspace{-0.2cm}
\resizebox{0.9\linewidth}{!}{%
\begin{tabularx}{\linewidth}{l  C  C  C  C  C }
\toprule
\footnotesize{Backbone} & \footnotesize$P_{cnn}$ & \footnotesize$V'$ & \footnotesize$C$ & \footnotesize{Error} & \footnotesize$P_{total}$
\\ 
 \midrule
\footnotesize{ResNet-50} & \footnotesize23.5M & \footnotesize804 & \footnotesize256 & \footnotesize1.38 & \footnotesize58.3M  \\
\footnotesize{EfficientNet-B0} & \footnotesize5.3M & \footnotesize160 & \footnotesize256 & \footnotesize4.12 & \footnotesize42.8M  \\
\footnotesize{EfficientNet-B0} & \footnotesize5.3M & \footnotesize160 & \footnotesize128 & \footnotesize4.96 & \footnotesize37.8M  \\
\footnotesize{EfficientNet-B0} & \footnotesize5.3M & \footnotesize80 & \footnotesize64 & \footnotesize\textbf{6.89} & \footnotesize\textbf{34.2M}  \\
\bottomrule
\end{tabularx}
}
\end{center}
\vspace{-0.4cm}
\end{table} 

%% file: fig/qualitative/handbooth/item.tex
\begin{figure}
\centering
% \includesvg[width=0.85\linewidth]{fig/qualitative/handbooth/handbooth_1.svg}
\includegraphics[width=0.8\linewidth]{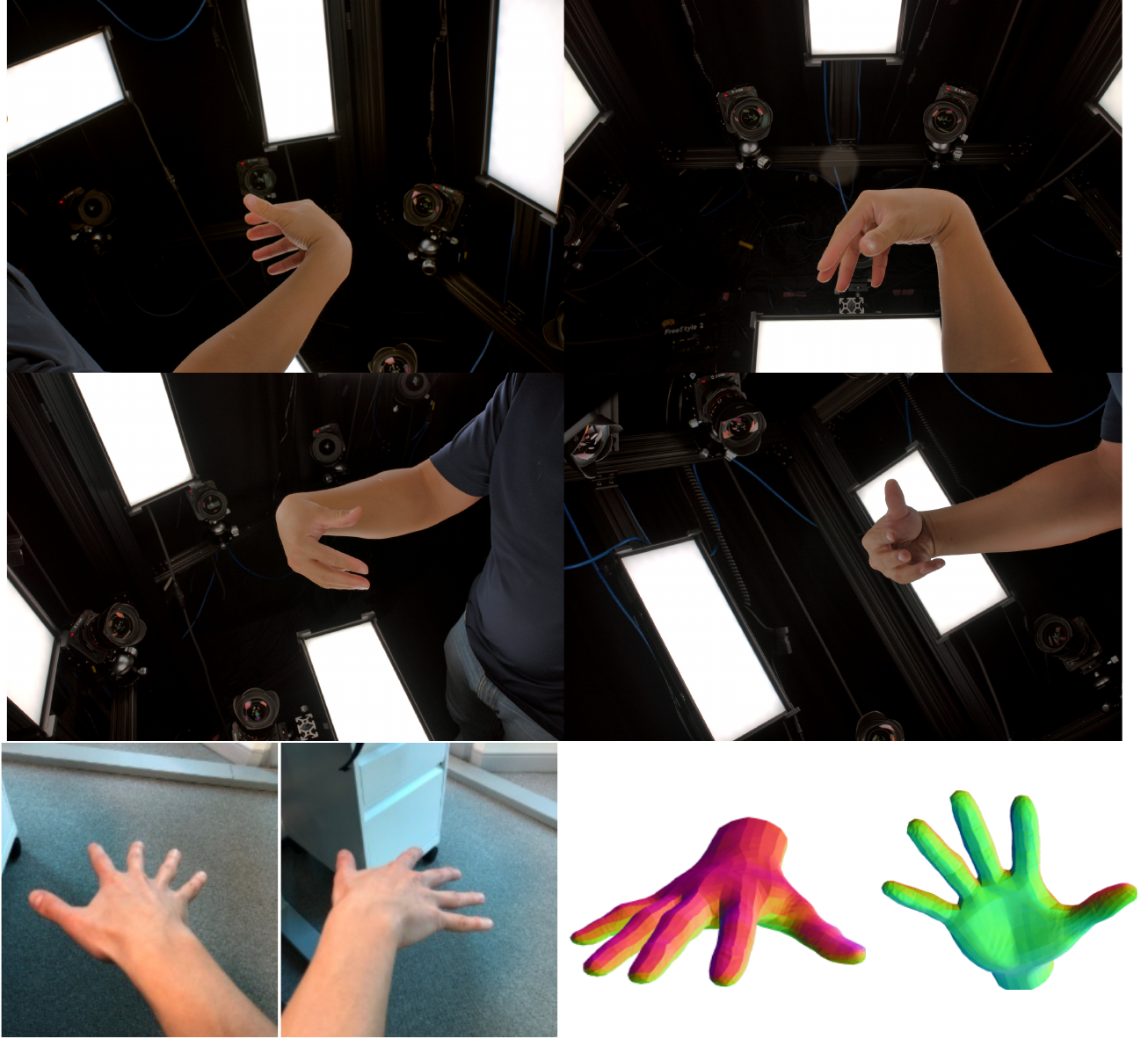}
\caption{Training examples of our real dataset (\emph{top}) and $3$D hand mesh estimation results on in-the-wild image (\emph{bottom}). The resulting mesh contains $3745$ vertices.
}
\vspace{-0.4cm}
\label{fig:handbooth}
\end{figure}

%% file: tex/conclusion.tex
\vspace{-0.1cm}
\section{Conclusion}
\vspace{-0.1cm}
In this paper, we have proposed a novel spectral graph-based transformer framework which reconstructs high-fidelity two-hand meshes from egocentric views. The main idea behind this study was to demonstrate that the fundamental properties of the graph Laplacian from the spectral graph theory can be applied to a Transformer. 
We have shown that our multi-view feature fusion strategy is most effective when coupled with mesh segmentation using spectral clustering. 
We further present spectral filtering and an optimisation-based refinement to reconstruct more physically plausible meshes. 
Our framework is general and can be extended to other multi-view reconstruction tasks. 
\vspace{-0.1cm}

% \noindent\emph{Limitations.} 
%1. learnable Laplacian or estimate on the fly

%% file: fig/qualitative/collision_full/item.tex
\begin{figure*}
\centering
\includegraphics[width=1\linewidth]{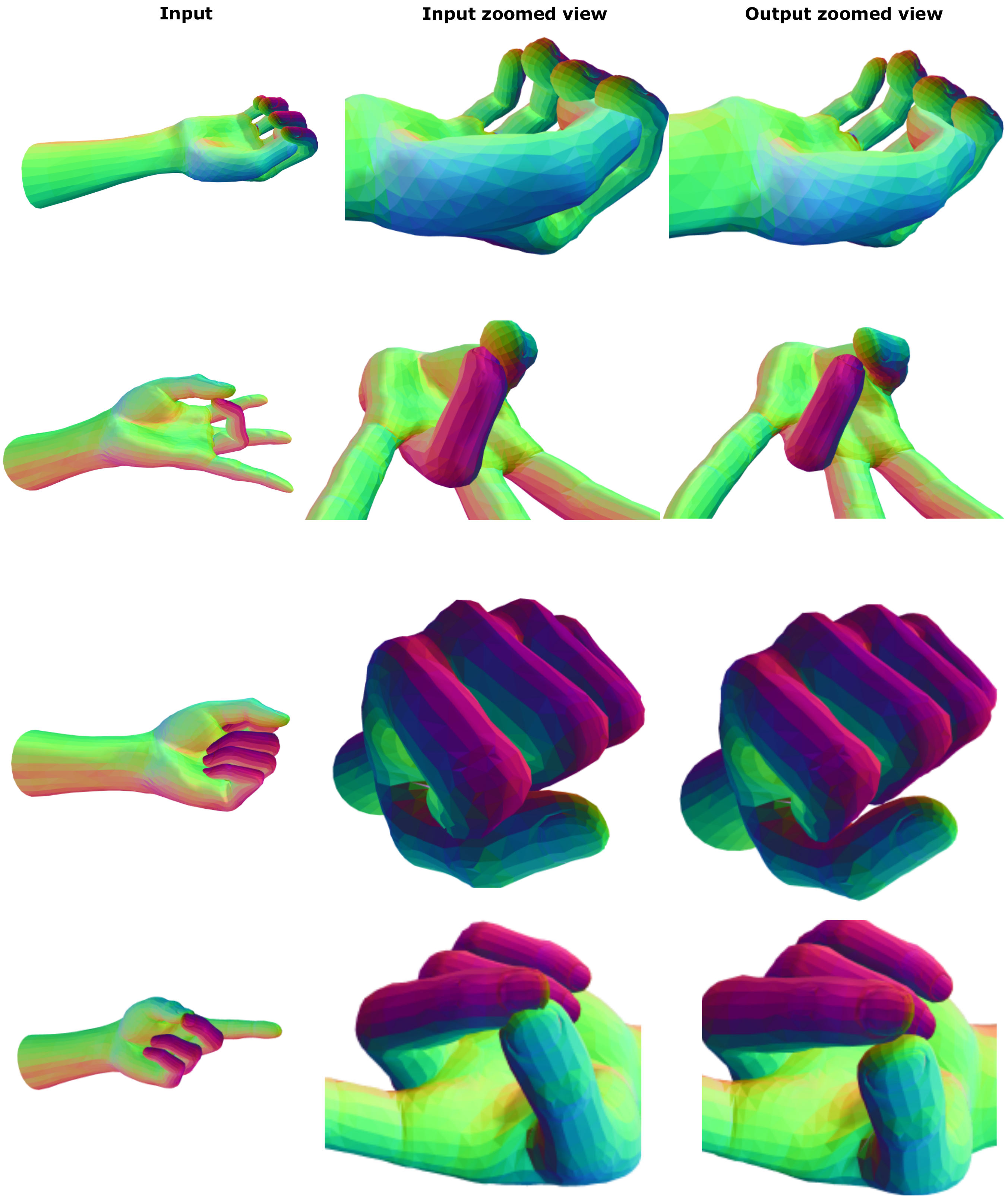}
\caption{Additional qualitative examples of mesh refinement at inference. Our optimisation-based strategy shows robustness to various hand poses. We present failure cases when self-penetration is highly complex in the supplementary materials.
}
\label{fig:mesh_refinement_full}
\end{figure*} 

%% file: tex/supplementary.tex
In this supplemental document, we provide:
% \begin{itemize}
%     \setlength{\itemsep}{0pt}%
%     \setlength{\parskip}{0pt}%
%     \item a summary of existing RGB hand pose estimation benchmarks (Sec \ref{sec:datasets});
%     \item details of our proposed network architectures, along with two baseline methods (Sec \ref{sec:network});
%     \item implementation details of spectral filters (Sec \ref{sec:filters});
%     \item analysis of different loss terms (Sec \ref{sec:loss_terms});
%     \item analysis on mesh refinement (Sec \ref{sec:mesh_refinement});
%     \item complete results for model compression (Sec \ref{sec:model_compression});
%     \item details of different fusion strategies (Sec \ref{sec:fusion});
%     \item additional $3$D mesh reconstruction results on our real dataset (Sec \ref{sec:real_data});
%     \item limitations of our method (Sec \ref{sec:limitations});
%     \item additional qualitative examples (Sec \ref{sec:examples}).
% \end{itemize}
\begin{itemize}
    \setlength{\itemsep}{0pt}%
    \setlength{\parskip}{0pt}%
    \item a summary of existing RGB hand pose estimation benchmarks (Sec \ref{sec:datasets});
    \item implementation details for networks and spectral filters (Sec \ref{sec:implementation});
    \item additional results and analysis (Sec \ref{sec:additional_analysis});
    \item limitations of our method (Sec \ref{sec:limitations});
    \item additional qualitative examples (Sec \ref{sec:examples}).
\end{itemize}

\section{Overview of existing datasets} \label{sec:datasets}
Table \ref{table:sup_dataset} presents a summary of existing datasets for hand pose estimation. As existing egocentric datasets are either instrumented with visible markers \cite{garcia2018first} or having limited background variation \cite{kwon2021h2o}, both of our synthetic and real datasets contain densely annotated two-hand with forearms.   
\input{tbl/supplementary/dataset}

\section{Implementation details}\label{sec:implementation}
In this section, we detail the network architectures in Section \ref{sec:network} and provide implementation details of spectral filters in Section \ref{sec:filters}.
\subsection{Network architectures} \label{sec:network}
\paragraph{Our proposed method.} We provide in Table \ref{table:sup_network_our} the full details of our network architecture. We use ResNet-$50$ \cite{he2016deep} as the backbone of our network. The input to our network model is $N$ number of multi-view $224 \times 224 \times 3$ RGB images and the output are two hand meshes with $4023$ vertices each hand.
\input{tbl/supplementary/network_ours}

\paragraph{METRO baseline.} We extend METRO \cite{lin2021end} from single-view setting to multi-view and detail the network architecture in Table \ref{table:sup_network_METRO}. Each transformer encoder block has $4$ layers and $3$ attention heads. Other than the size of feature dimension, the hyperparameters for transformer encoders are consistent across our proposed method and this baseline.
\input{tbl/supplementary/network_METRO}

\paragraph{Parametric baseline.} We detail the network architecture of our parametric baseline in Table \ref{table:sup_network_parametric}. The output of the network is passed to a pre-computed parametric hand model to produce hand mesh with $4023$ vertices.
\input{tbl/supplementary/network_parametric}

\subsection{Implementation details of spectral filters} \label{sec:filters}
We perform spectral filtering by applying a customised filter to the spectral representation of the graph signal. The spectral representation depends on the adjacency matrix and the eigenvectors of the graph. Spectral filtering is performed by applying a filter function to the spectral coefficients. In the following, we detail Gaussian and Laplacian filters. 

Given eigenvalue $\lambda$, the Gaussian filter function $f_{gau}$ can be described as:
\vspace{-0.1cm}
\begin{align} 
    f_{gau}(\lambda) = e^{\frac{-\lambda^2}{2\sigma^2}},
\end{align}
where $\sigma$ is the standard deviation of the Gaussian distribution. We set $\sigma=0.5$ in our experiments.
Similarly, the Laplacian filter function $f_{lap}$ can be described as:
\begin{align} 
    f_{lap}(\lambda) = \lambda^{-\frac{1}{2}}.
\end{align}
As eigenvalues can have zero and negative values, the inverse square root computation can result in NaN (not a number) values. Therefore, we put a tolerance value that is close to zero to avoid this.

\section{Additional analysis}\label{sec:additional_analysis}
\subsection{Influence of different loss terms} \label{sec:loss_terms}
In Table \ref{table:sup_loss}, we analyse the influence of different loss terms. In these experiments, we consider L$1$ losses on $3$D mesh vertices and $2$D re-projection loss, \ie $\loss_{mesh}$ and $\loss_{2D}$, respectively. In addition, we experiment with mean squared euclidean distance loss $\loss_{MSE}$ on hand mesh. For mesh regularisation, we apply edge length regularisation $\loss_{edge}$ and minimise the Chamfer distances $\loss_{cham}$. We find that the combination of $\loss_{mesh}$, $\loss_{2D}$ and $\loss_{edge}$ delivers the optimal results.
\input{tbl/supplementary/loss}

\subsection{Analysis on mesh refinement at inference} \label{sec:mesh_refinement}
For quantitative evaluation, we followed \cite{hasson2019learning, tse2022collaborative, tse2022s} to include penetration depth $(mm)$ and intersection volume $(cm^3)$. Penetration depth refers to the maximum distances from hand mesh vertices to the other/self hand’s surface when in a collision. Intersection volume is obtained by voxelising the meshes using a voxel size of $0.5cm$. We report the results in Table \ref{table:sup_mesh_refinement}. 
% We provide additional qualitative examples in \fig{sup_mesh_refinement}. 
These results demonstrate the robustness of our optimisation-based mesh refinement strategy. 

\input{tbl/supplementary/mesh_refinement}

\subsection{Complete results for model compression}\label{sec:model_compression}
We provide complete experimental results for model compression in Table \ref{table:sup_smaller_model}. We find that the feature channel size $C$ cannot drop under $64$ as performance drops massively. Therefore, we fix $C=64$ and reduce the number of vertices for template hand mesh $V'$. 

\input{tbl/supplementary/model_compression}

\subsection{Complete results for multi-view fusion}
We provide complete experimental results for different multi-view fusion strategies in Table \ref{table:full_abl_multi_view}. We find that the
combination of soft-attention fusion and mesh segmentation
consistently improves the performance.

\input{tbl/supplementary/full_multi_view}

\subsection{Illustration of different fusion strategies} \label{sec:fusion}
We illustrate two fusion strategies in \fig{sup_fusion} which were considered for Table 3 in the main text. In \fig{sup_fusion}a), as max pooling is applied on each of the input image separately, the resulting features are in size of $[N \times 2048]$. On the other hand, in \fig{sup_fusion}b), only one global feature vector is computed per data sample by applying max pool across multi-view input images.

\subsection{Additional quantitative comparison on real dataset} \label{sec:real_data}
We provide additional quantitative comparison with METRO \cite{lin2021end} using our real dataset in Table \ref{table:sup_main_real}. In these experiments, we vary the number of input camera views for training and withheld $3$ unseen camera views for evaluation. Our method consistently outperforms our METRO baseline across different number of input views. This shows the generalisation ability of our proposed method.

\input{tbl/supplementary/main_real}

\section{Method limitations}\label{sec:limitations}
Though our method results in accurate and physically-plausible high fidelity two-hand reconstructions, the results are sometimes not plausible when self-penetration is highly complex (see \fig{sup_refinment_fail}). We believe this problem can be tackled in the future by incorporating temporal information and more advance physical modeling into our proposed framework. In particular, the current optimisation-based mesh refinement works on self-penetrations only. Solving interpenetrations during hand-hand interactions is in-line with our future goal.

\section{Additional examples}\label{sec:examples}
We provide additional qualitative examples on our synthetic dataset and real dataset in \fig{sup_synthetic} and \fig{sup_real}, respectively. 

\input{fig/supplementary/mesh_refinement_fail/item}
\input{fig/supplementary/fusion/item}
\input{fig/supplementary/synthetic/item}
\input{fig/supplementary/handbooth/item}

%% file: tbl/supplementary/dataset.tex
\newcolumntype{C}{>{\centering\arraybackslash}X}
\begin{table*}
\begin{center}
\caption{A comparison of existing RGB hand pose estimation benchmarks.
}
\label{table:sup_dataset}
% \vspace{-0.2cm}
\resizebox{0.9\linewidth}{!}{%
\begin{tabularx}{\linewidth}{l | C  C  C  C  C  C}
\toprule
 & \# Frames & Two-hand & Egocentric  & Markerless & Background variation & Dense annotation \\
 \midrule
FPHA \cite{garcia2018first} & $105$k & \xmark &  \cmark &  \xmark & \xmark & \xmark \\
FreiHAND \cite{zimmermann2019freihand} & $37$k & \xmark & \xmark &  \cmark & \xmark & \xmark\\
InterHand$2.6$M \cite{moon2020interhand2} & $2.6$M & \cmark & \xmark &  \cmark & \xmark & \xmark \\
H$2$O \cite{kwon2021h2o} & $571$k & \cmark & \cmark &  \cmark & \xmark & \xmark\\
H$_{2}$O-$3$D \cite{hampali2022keypoint} & $76$k & \cmark & \xmark & \cmark & \xmark & \xmark\\
Ego$3$DHands \cite{lin2021two} & $55$k & \cmark & \cmark & \cmark & \cmark & \xmark\\
\bottomrule
Ours (synthetic) & $1$M & \cmark & \cmark & \cmark & \cmark & \cmark\\
Ours (real) & $61$k & \cmark & \cmark & \cmark & \xmark & \cmark\\
\bottomrule
\end{tabularx}
}
\end{center}
% \vspace{-0.4cm}
\end{table*} 

%% file: tbl/supplementary/network_ours.tex
\newcolumntype{C}{>{\centering\arraybackslash}X}
\begin{table}
\begin{center}
%\vspace{-0.2cm}
\caption{Architecture of our network. $B$ refers to batch size and $N$ refers to the number of multi-view RGB input images. Note that the duplicated layer numbers are performed in parallel and the output of layer $13$ is concatenated with template hand mesh $804 \times 3$ before feeding to transformer encoder in layer $14$.
}
\label{table:sup_network_our}
\vspace{-0.2cm}
\resizebox{0.9\linewidth}{!}{%
\begin{tabularx}{\linewidth}{l|  C  |C }
\toprule
Layer & Operation & Dimensionality \\
\midrule 
 & Input & $B \times N \times 224 \times 224 \times 3$ \\
\midrule 
1 & ResNet-$50$ & $B \times N  \times 7 \times 7 \times 2048$ \\
\midrule 
2 & Upsampling $2$D & $B \times N  \times 14 \times 14 \times 2048$ \\
3 & Convolution $2$D & $B \times N  \times 12 \times 12 \times 256$ \\
4 & Batch normalisation & $B \times N  \times 12 \times 12 \times 256$ \\
5 & ReLU & $B \times N  \times 12 \times 12 \times 256$ \\
6 & Upsampling $2$D & $B \times N  \times 24 \times 24 \times 256$ \\
7 & Convolution $2$D & $B \times N  \times 22 \times 22 \times 256$ \\
8 & Batch normalisation & $B \times N  \times 22 \times 22 \times 256$ \\
9 & ReLU & $B \times N  \times 22 \times 22 \times 256$ \\
\midrule 
2 & Upsampling $2$D & $B \times N  \times 14 \times 14 \times 2048$ \\
3 & Convolution $2$D & $B \times N  \times 12 \times 12 \times 256$ \\
4 & Batch normalisation & $B \times N  \times 12 \times 12 \times 256$ \\
5 & ReLU & $B \times N  \times 12 \times 12 \times 256$ \\
6 & Upsampling $2$D & $B \times N  \times 24 \times 24 \times 256$ \\
7 & Convolution $2$D & $B \times N  \times 22 \times 22 \times 256$ \\
8 & Batch normalisation & $B \times N  \times 22 \times 22 \times 256$ \\
9 & ReLU & $B \times N  \times 22 \times 22 \times 256$ \\
10 & Convolution $1$D & $B \times N  \times 22 \times 22 \times 7$\\
11 & Spatial softmax & $B \times N  \times 22 \times 22 \times 7$ \\
\midrule
12 & Matrix multiplication & $B \times N  \times 7 \times 256$ \\
13 & Max pooling & $B \times 7 \times 256$ \\
\midrule
14 & Transformer encoder & $B \times 804 \times 259$ \\
15 & Fully-connected layer & $B \times 804 \times 130$ \\
16 & Transformer encoder & $B \times 804 \times 130$ \\
17 & Fully-connected layer & $B \times 804 \times 65$ \\
18 & Transformer encoder & $B \times 804 \times 65$ \\
19 & Fully-connected layer & $B \times 804 \times 3$ \\
\midrule
20 & Fully-connected layer & $B \times 617 \times 3$\\
21 & Spectral filtering & $B \times 617 \times 3$ \\
22 & Fully-connected layer & $B \times 1234 \times 3$\\
23 & Spectral filtering & $B \times 1234 \times 3$ \\
24 & Fully-connected layer & $B \times 2468 \times 3$\\
25 & Spectral filtering & $B \times 2468 \times 3$ \\
26 & Fully-connected layer & $B \times 4023 \times 3$ \\
\midrule
20 & Fully-connected layer & $B \times 617 \times 3$\\
21 & Spectral filtering & $B \times 617 \times 3$ \\
22 & Fully-connected layer & $B \times 1234 \times 3$\\
23 & Spectral filtering & $B \times 1234 \times 3$ \\
24 & Fully-connected layer & $B \times 2468 \times 3$\\
25 & Spectral filtering & $B \times 2468 \times 3$ \\
26 & Fully-connected layer & $B \times 4023 \times 3$ \\
\midrule
 & Output & $B \times 8046 \times 3$ \\
\bottomrule
\end{tabularx}
}
%\vspace{-0.4cm}
\end{center}
\end{table}

%% file: tbl/supplementary/network_METRO.tex
\newcolumntype{C}{>{\centering\arraybackslash}X}
\begin{table}
\begin{center}
%\vspace{-0.2cm}
\caption{Architecture of METRO baseline.
}
\label{table:sup_network_METRO}
\vspace{-0.2cm}
\resizebox{0.9\linewidth}{!}{%
\begin{tabularx}{\linewidth}{l|  C  |C }
\toprule
Layer & Operation & Dimensionality \\
\midrule 
 & Input & $B \times N \times 224 \times 224 \times 3$ \\
\midrule 
1 & ResNet-$50$ & $B \times N  \times 7 \times 7 \times 2048$ \\
2 & Max pooling $2$D & $B \times 2048$ \\
\midrule 
3 & Transformer encoder & $B \times 804 \times 2051$ \\
4 & Fully-connected layer & $B \times 804 \times 1026$ \\
5 & Transformer encoder & $B \times 804 \times 1026$ \\
6 & Fully-connected layer & $B \times 804 \times 513$ \\
7 & Transformer encoder & $B \times 804 \times 513$ \\
8 & Fully-connected layer & $B \times 804 \times 256$ \\
9 & Fully-connected layer & $B \times 804 \times 3$ \\
\midrule
10 & Fully-connected layer & $B \times 1624 \times 3$\\
11 & Fully-connected layer & $B \times 4023 \times 3$\\
12 & Fully-connected layer & $B \times 8046 \times 3$\\
\midrule
 & Output & $B \times 8046 \times 3$ \\
\bottomrule
\end{tabularx}
}
%\vspace{-0.4cm}
\end{center}
\end{table}

%% file: tbl/supplementary/network_parametric.tex
\newcolumntype{C}{>{\centering\arraybackslash}X}
\begin{table}
\begin{center}
%\vspace{-0.2cm}
\caption{Architecture of parametric baseline.
}
\label{table:sup_network_parametric}
\vspace{-0.2cm}
\resizebox{0.9\linewidth}{!}{%
\begin{tabularx}{\linewidth}{l|  C  |C }
\toprule
Layer & Operation & Dimensionality \\
\midrule 
 & Input & $B \times N \times 224 \times 224 \times 3$ \\
\midrule 
1 & ResNet-$50$ & $B \times N  \times 7 \times 7 \times 2048$ \\
2 & Max pooling $2$D & $B \times 2048$ \\
\midrule
3 & Fully-connected layer & $B \times 1024$\\
4 & Fully-connected layer & $B \times 512$\\
5 & Fully-connected layer & $B \times 200$\\
\midrule
 & Output & $B \times 200$ \\
\bottomrule
\end{tabularx}
}
%\vspace{-0.4cm}
\end{center}
\end{table}

%% file: tbl/supplementary/loss.tex
\newcolumntype{C}{>{\centering\arraybackslash}X}
\begin{table}
\begin{center}
\caption{Impact of different loss terms on our synthetic dataset. Hand errors are given in millimeters (mm). 
}
\label{table:sup_loss}
% \vspace{-0.2cm}
\resizebox{0.9\linewidth}{!}{%
\begin{tabularx}{\linewidth}{C  C C C C| C }
\toprule
$\loss_{mesh}$ & $\loss_{2D}$ & $\loss_{MSE}$ & $\loss_{edge}$ & $\loss_{cham}$ & Error  \\ 
\midrule 
\cmark & \cmark & & & & 1.55\\
\cmark & \cmark & \cmark & & & 1.48\\
\cmark & \cmark & & \cmark & & 1.38\\
\cmark & \cmark & \cmark & \cmark & & 1.38\\
\cmark & \cmark & \cmark & \cmark & \cmark & 1.49\\
\bottomrule
\end{tabularx}
}
\end{center}
\vspace{-0.4cm}
\end{table}

%% file: tbl/supplementary/mesh_refinement.tex
\newcolumntype{C}{>{\centering\arraybackslash}X}
\begin{table}
\begin{center}
%\vspace{-0.2cm}
\caption{Quantitative evaluation on the impact of mesh refinement at inference.
}
\label{table:sup_mesh_refinement}
\vspace{-0.2cm}
\resizebox{0.9\linewidth}{!}{%
\begin{tabularx}{\linewidth}{l|  C  |C }
\toprule
 & Before refinement & After refinement \\
\midrule 
Max. penetration ($mm$) & $5.3$ & $0.16$ \\
Intersection vol. ($ cm^3$) & $2.0$ & $0.09$ \\
\bottomrule
\end{tabularx}
}
%\vspace{-0.4cm}
\end{center}
\end{table}

%% file: tbl/supplementary/model_compression.tex
\newcolumntype{C}{>{\centering\arraybackslash}X}
\begin{table}
\begin{center}
\caption{Ablations of different backbones and hyperparameters. We denote $P_{cnn}$ and $P_{total}$ to be the number of parameters for CNN backbone and total model, respectively.
}
\label{table:sup_smaller_model}
% \vspace{-0.2cm}
\resizebox{1\linewidth}{!}{%
\begin{tabularx}{\linewidth}{l  C  C  C  C  C }
\toprule
\footnotesize{Backbone} & \footnotesize$P_{cnn}$ & \footnotesize$V'$ & \footnotesize$C$ & \footnotesize{Error} & \footnotesize$P_{total}$
\\ 
 \midrule
\footnotesize{ResNet-50} & \footnotesize23.5M & \footnotesize804 & \footnotesize256 & \footnotesize1.38 & \footnotesize58.3M  \\
 \midrule
 \footnotesize{EfficientNet-B3} & \footnotesize12.9M & \footnotesize804 & \footnotesize256 & \footnotesize2.53 & \footnotesize47.7M  \\
\footnotesize{EfficientNet-B2} & \footnotesize9.2M & \footnotesize804 & \footnotesize256 & \footnotesize2.55 & \footnotesize44M  \\
\footnotesize{EfficientNet-B1} & \footnotesize7.8M & \footnotesize804 & \footnotesize256 & \footnotesize2.74 & \footnotesize42.6M  \\
\footnotesize{EfficientNet-B0} & \footnotesize5.3M & \footnotesize804 & \footnotesize256 & \footnotesize2.85 & \footnotesize40.1M  \\
 \midrule
 \footnotesize{EfficientNet-B3} & \footnotesize12.9M & \footnotesize804 & \footnotesize128 & \footnotesize3.00 & \footnotesize40.6M  \\
\footnotesize{EfficientNet-B2} & \footnotesize9.2M & \footnotesize804 & \footnotesize128 & \footnotesize3.20 & \footnotesize36.9M  \\
\footnotesize{EfficientNet-B1} & \footnotesize7.8M & \footnotesize804 & \footnotesize128 & \footnotesize3.20 & \footnotesize35.5M  \\
\footnotesize{EfficientNet-B0} & \footnotesize5.3M & \footnotesize804 & \footnotesize128 & \footnotesize2.91 & \footnotesize33M  \\
 \midrule
  \footnotesize{EfficientNet-B3} & \footnotesize12.9M & \footnotesize804 & \footnotesize64 & \footnotesize4.00 & \footnotesize38.4M  \\
\footnotesize{EfficientNet-B2} & \footnotesize9.2M & \footnotesize804 & \footnotesize64 & \footnotesize3.87 & \footnotesize34.7M  \\
\footnotesize{EfficientNet-B1} & \footnotesize7.8M & \footnotesize804 & \footnotesize64 & \footnotesize4.04 & \footnotesize33.3M  \\
\footnotesize{EfficientNet-B0} & \footnotesize5.3M & \footnotesize804 & \footnotesize64 & \footnotesize4.31 & \footnotesize30.8M  \\
 \midrule
 \footnotesize{EfficientNet-B3} & \footnotesize12.9M & \footnotesize804 & \footnotesize32 & \footnotesize89.7 & \footnotesize37.3M  \\
\footnotesize{EfficientNet-B0} & \footnotesize9.2M & \footnotesize804 & \footnotesize32 & \footnotesize90 & \footnotesize29.7M  \\
 \midrule
\footnotesize{EfficientNet-B0} & \footnotesize5.3M & \footnotesize160 & \footnotesize256 & \footnotesize4.12 & \footnotesize42.8M  \\
\footnotesize{EfficientNet-B0} & \footnotesize5.3M & \footnotesize160 & \footnotesize128 & \footnotesize4.96 & \footnotesize37.8M  \\
\footnotesize{EfficientNet-B0} & \footnotesize5.3M & \footnotesize80 & \footnotesize64 & \footnotesize6.89 & \footnotesize34.2M \\
\bottomrule
\end{tabularx}
}
\end{center}
\vspace{-0.2cm}
\end{table} 

%% file: tbl/supplementary/full_multi_view.tex
\newcolumntype{C}{>{\centering\arraybackslash}X}
\begin{table}
\begin{center}
\caption{Performance of different multi-view fusion strategies. We report hand error for both settings. $K$ refers to the number of clusters for template hand mesh. Note that we do not include spectral filtering in the graph decoder here.
}
\label{table:full_abl_multi_view}
% \vspace{-0.2cm}
\resizebox{0.9\linewidth}{!}{%
\begin{tabularx}{\linewidth}{l | C | C }
\toprule
 & Single-view & Multi-view \\ 
\midrule 
METRO \cite{lin2021end} & 10.87 & - \\
METRO \cite{lin2021end} + avg. pool & - & 8.71 \\ 
METRO \cite{lin2021end} + max pool & - & 7.09 \\ 
\midrule
Ours ($K=1$) & - & 6.59\\
Ours ($K=2$) & - & 5.71\\
Ours ($K=3$) & - & 5.19\\
Ours ($K=4$) & - & 4.79\\
Ours ($K=5$) & - & 4.59\\
Ours ($K=6$) & - & 5.28\\
Ours ($K=7$) & - & \textbf{3.72} \\
Ours ($K=8$) & - & 3.79\\
\bottomrule
\end{tabularx}
}
\end{center}
\vspace{-0.4cm}
\end{table}

%% file: tbl/supplementary/main_real.tex
\newcolumntype{C}{>{\centering\arraybackslash}X}
\begin{table}
\begin{center}
\caption{Error rates on our real dataset. Our method consistently outperforms METRO on different number of input views.
}
\label{table:sup_main_real}
% \vspace{-0.2cm}
\resizebox{1\linewidth}{!}{%
\begin{tabularx}{\linewidth}{l | C  C  C}
\toprule
 & 5-views & 3-views & 2-views  \\ 
 \midrule
 METRO \cite{lin2021end} & 9.67 & 12.0 & 16.1  \\
Ours & \textbf{4.34} & \textbf{6.94} & \textbf{7.73}  \\
\bottomrule
\end{tabularx}
}
\end{center}
\vspace{-0.4cm}
\end{table} 

%% file: fig/supplementary/mesh_refinement_fail/item.tex
\begin{figure}
\centering
\includegraphics[width=1\linewidth]{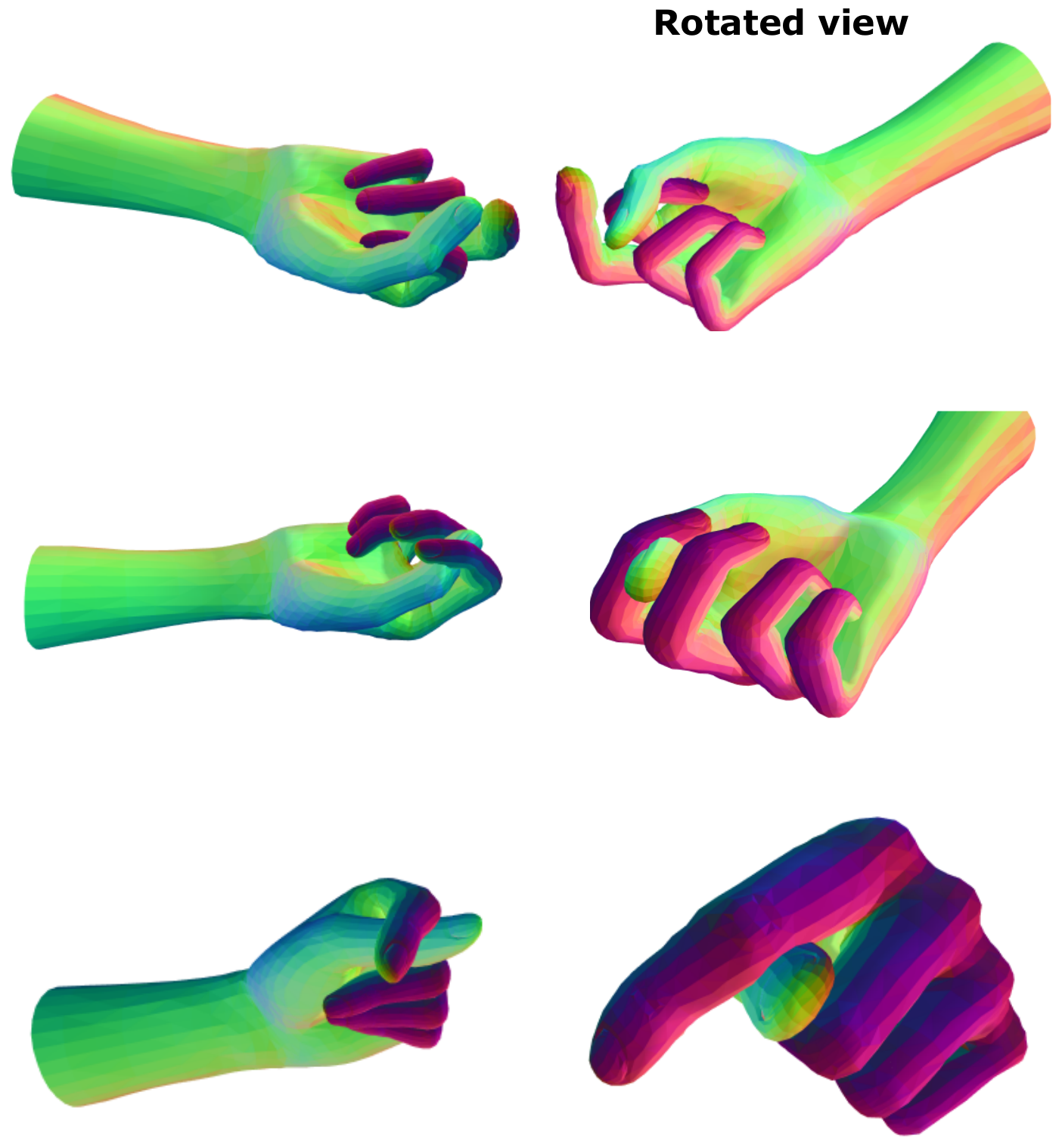}
\caption{Failure examples for mesh refinement.}
\label{fig:sup_refinment_fail}
\end{figure}                                 

%% file: fig/supplementary/fusion/item.tex
\begin{figure}
\centering
\includegraphics[width=1\linewidth]{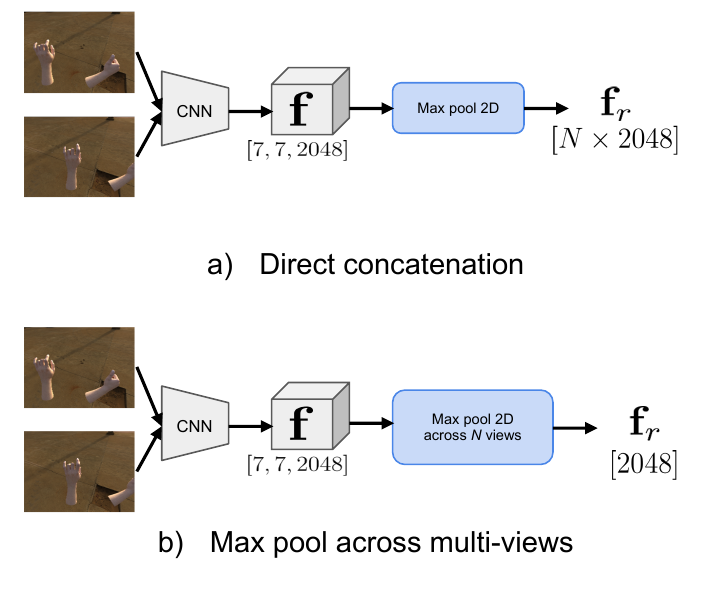}
\caption{Illustration of different feature fusion strategies.}
\label{fig:sup_fusion}
\end{figure}    
 

%% file: fig/supplementary/synthetic/item.tex
\begin{figure*}
\centering
\includegraphics[width=1\linewidth]{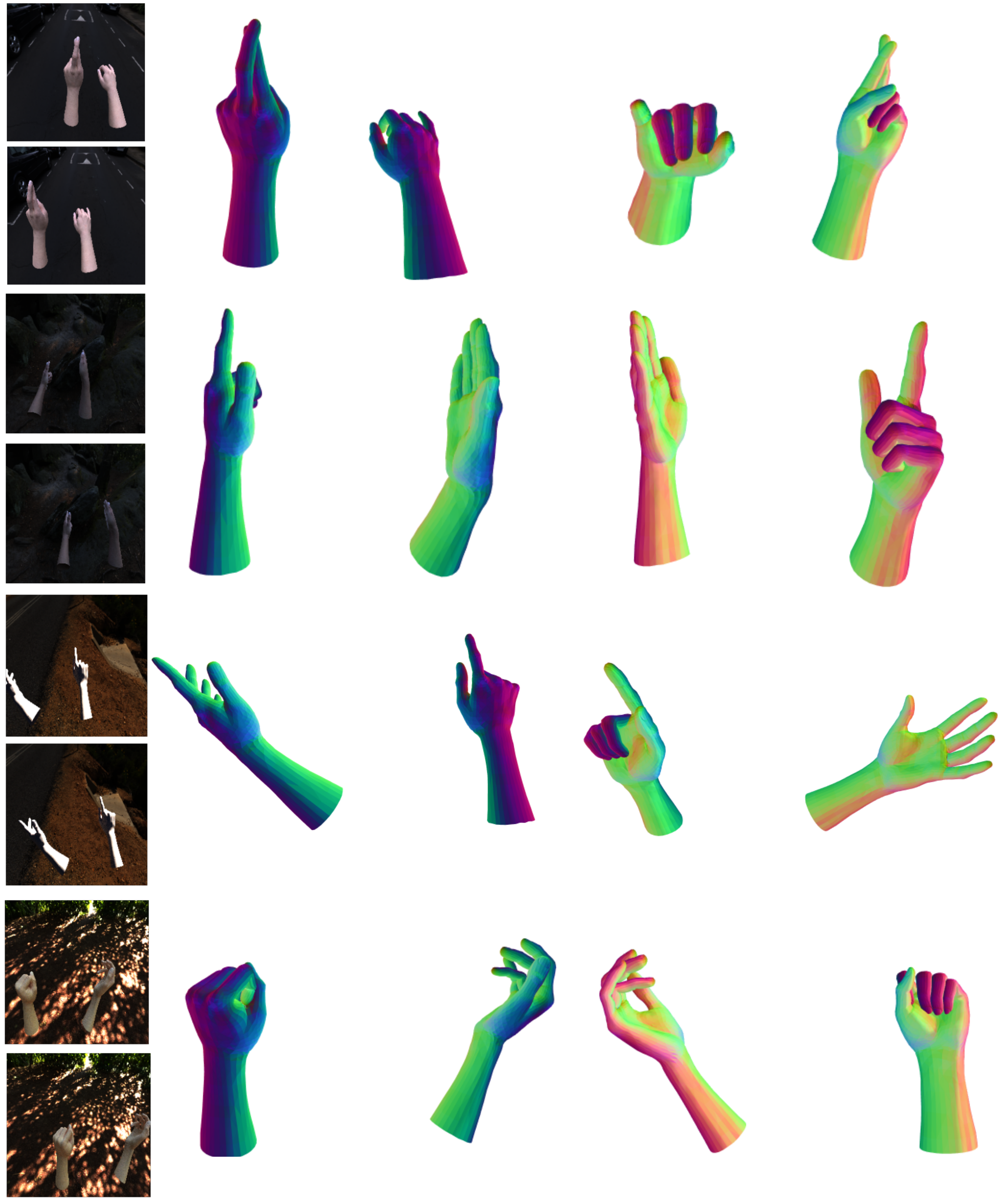}

\caption{Additional qualitative examples on our synthetic dataset.
}

\label{fig:sup_synthetic}
\end{figure*} 

%% file: fig/supplementary/handbooth/item.tex
\begin{figure*}
\centering
\includegraphics[width=1\linewidth]{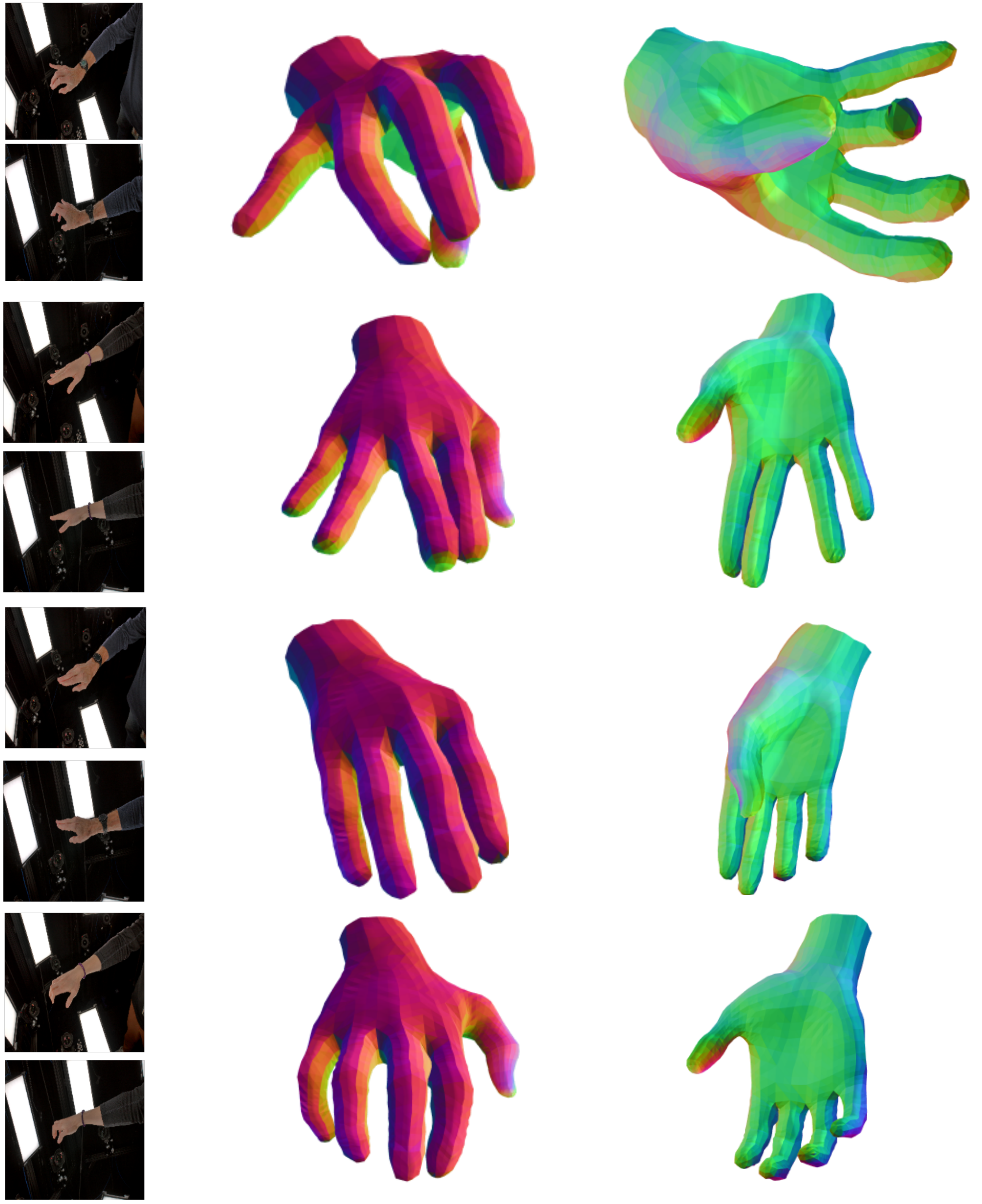}

\caption{Additional qualitative examples on our real dataset.}

\label{fig:sup_real}
\end{figure*} 